\documentclass{article}

\usepackage[final]{nips_2016}

\usepackage[utf8]{inputenc} 
\usepackage[T1]{fontenc}    
\usepackage{hyperref}       
\usepackage{url}            
\usepackage{booktabs}       
\usepackage{amsfonts}       
\usepackage{nicefrac}       
\usepackage{microtype}      

\usepackage{amssymb}
\usepackage{amsmath}
\usepackage{graphicx}
\usepackage{subfigure} 
\usepackage{algorithm}
\usepackage{algorithmic}
\usepackage[multiple]{footmisc}
\DeclareMathOperator{\tr}{tr}
\usepackage{url}
\usepackage{verbatim}
\usepackage{enumerate}
\usepackage{color}
\usepackage{bm}
\PassOptionsToPackage{square,sort,comma,numbers}{natbib}
\usepackage{booktabs}

\newsavebox\myboxA
\newsavebox\myboxB
\newlength\mylenA

\makeatletter
\newcommand*\xoverline[2][0.75]{%
    \sbox{\myboxA}{$\m@th#2$}%
    \setbox\myboxB\null
    \ht\myboxB=\ht\myboxA%
    \dp\myboxB=\dp\myboxA%
    \wd\myboxB=#1\wd\myboxA
    \sbox\myboxB{$\m@th\overline{\copy\myboxB}$}
    \setlength\mylenA{\the\wd\myboxA}
    \addtolength\mylenA{-\the\wd\myboxB}%
    \ifdim\wd\myboxB<\wd\myboxA%
       \rlap{\hskip 0.5\mylenA\usebox\myboxB}{\usebox\myboxA}%
    \else
        \hskip -0.5\mylenA\rlap{\usebox\myboxA}{\hskip 0.5\mylenA\usebox\myboxB}%
    \fi}
    \makeatletter
    
\newcommand{\D}{\ensuremath{\mathcal{D}}}
\newcommand{\X}{\ensuremath{\mathcal{X}}}
\newcommand{\C}{\ensuremath{\mathcal{C}}}
\newcommand{\kernel}{\ensuremath{\mathbf{K}}}
\newcommand{\Crel}{\ensuremath{\C_{\operatorname{neq}}}}
\newcommand{\Cdk}{\ensuremath{\C_{\operatorname{eq}}}}
\newcommand{\dfun}{\ensuremath{\delta}}
\newcommand{\kspace}{\ensuremath{\mathbb{R}^m}}
\newcommand{\transpose}{\ensuremath{^\top}}
\newcommand{\e}{\ensuremath{\mathbf{e}}}
\newcommand{\ratio}{\ensuremath{\gamma}}
\newcommand{\Cmat}{\ensuremath{\mathbf{C}}}

\newcommand{\divergence}{\ensuremath{\operatorname{D}}}
\newcommand{\logdet}{\ensuremath{\divergence_{\operatorname{ld}}}}
\newcommand{\trineq}[3]{\ensuremath{({#1},{#2}\mid{#3})}}
\newcommand{\trieq}[3]{\ensuremath{({#1},{#2},{#3})}}
\newcommand{\indneq}[3]{\ensuremath{\scriptscriptstyle{#1}{#2}{#3}}}
\newcommand{\indeq}[3]{\ensuremath{\xoverline[0.6]{\scriptscriptstyle{#1}{#2}{#3}}}}
\newcommand{\indneqbig}[3]{\ensuremath{{#1}{#2}{#3}}}
\newcommand{\indeqbig}[3]{\ensuremath{\xoverline[0.7]{{#1}{#2}{#3}}}}
\newcommand{\W}{\ensuremath{\mathbf{W}}}

\newcommand{\Iset}{\ensuremath{\mathcal{I}}}
\newcommand{\Eset}{\ensuremath{\mathcal{E}}}

\title{Semi-supervised Kernel Metric Learning\\Using Relative Comparisons}

\author{
Ehsan Amid\\
Department of Computer Science, UC - Santa Cruz \\
		Santa Cruz, CA, 95064\\
              \texttt{eamid@ucsc.edu} \\
\And
Aristides Gionis \\
Department of Computer Science, Aalto University\\
       Helsinki Institute for Information Technology\\
       02150 Espoo, Finland\\ \texttt{aristides.gionis@aalto.fi } \\
\AND
Antti Ukkonen \\
Finnish Institute of Occupational Health \\
       00290 Helsinki, Finland\\ \texttt{antti.ukkonen@ttl.fi }\\
}

\begin{document}

\maketitle

\begin{abstract}
We consider the problem of metric learning
subject to a set of constraints on relative-distance comparisons between the data items.
Such constraints are meant to  
reflect side-information that is not expressed directly in the feature vectors of the data items. 
The relative-distance constraints used in this work
are particularly effective in expressing structures at finer level of detail
than must-link (ML) and cannot-link (CL) constraints, 
which are most commonly used for semi-supervised clustering.
Relative-distance constraints are thus useful in settings
where providing an ML or a CL constraint
is difficult because the granularity of the true clustering is unknown.

Our main contribution is an efficient algorithm
for learning a kernel matrix using the log determinant divergence 
--- a variant of the Bregman divergence ---
subject to a set of relative-distance constraints.
The learned kernel matrix can then be employed by many different kernel methods in a wide range of applications.
In our experimental evaluations, we consider a semi-supervised clustering setting and
show empirically that kernels found by our algorithm
yield clusterings of higher quality than existing approaches
that either use ML/CL constraints
or a different means to
implement the supervision using relative comparisons.
\end{abstract}

\section{Introduction}
\label{sec:intro}

{\em Metric learning} is the task of finding an appropriate metric (distance function) between a set of items. 
In many cases, 
a feature representation of the items is provided as input 
and distances between the data items can be calculated using 
a proper norm on the corresponding feature vectors.
However, it is often the case that the feature-vector representation of the items alone 
is not sufficient to describe intricate and refined relations in the data. 
For instance,
when clustering images
it may be necessary to make use of
semantic information about the image content,
in addition to some standard image-processing features.

{\em Semi-supervised metric learning} provides a principled framework 
for combining feature vectors with other external information 
that can help capturing more refined relations in the data. 
Such external information is usually given as
labels about {\em pair-wise distances} between a few data items.
Such labels may be obtained by crowd-sourcing, 
or provided by the data analyst, 
or anyone interacting with the clustering application, 
and they reflect properties of the data that are not expressed directly from the data features.
The metric-learning problem then corresponds to finding a linear transformation of the initial features 
such that the constraints imposed by the external information are satisfied.

There are two commonly used ways to
formalize such side information.
The first
are {\em must-link} (ML) and {\em cannot-link} (CL) constraints.
An ML (CL) constraint between data items $i$ and $j$ suggests that the two
items
are similar (dissimilar), and
should thus be placed close to (far away from) each other.
These types of constraints can be generated using a subset of labeled items where an ML (CL) constraint represents a pair of items from the same class (different classes).
The second way to express pair-wise similarities
are {\em relative distance comparisons}.
These are statements that specify how
the distances between some data items relate to each other.
The most common relative distance comparison task
asks the data analyst to
specify which of the items $i$ and $j$
is closer to a third item $k$.
Note that unlike the ML/CL constraints,
the relative comparisons
do not as such say anything about
the clustering structure.

Given a number of distance constraints between few data items,
provided as examples, 
the objective of metric learning is to devise a new distance function, 
which takes into account both the supplied features and the additional distance constraints, 
and which expresses better the semantics of the application. 
Approaches using both types of constraints discussed above, 
ML/CL constraints and relative distance comparisons, 
have been studied in the literature, 
and a lot is known about the problem.

The method we discuss in this paper
is {\em a combination of metric-learning
and relative distance comparisons}.
We deviate from
existing literature
by eliciting
every constraint
with the question
\begin{quote}
\emph{``Given three items $i$, $j$, and $k$, which one is the least similar to the other two?''}
\end{quote}
The labeler should thus select one of the items as an {\em outlier}.
Such tasks have been used e.g., by \citet{crowdmedian} and \citet{UkkonenDH15}.
Notably,
we also allow the labeler to
leave the answer as {\em unspecified}.
The main practical novelty of
this approach is in the
{\em capability to gain information also from comparisons
where the labeler has
not been able to give a concrete solution}.
Some sets of three items can be
all very similar (or dissimilar) to each other,
so that picking one item as an obvious outlier
is difficult.
In those cases that the labeler gives a ``don't know'' answer,
it is beneficial to use this answer in the metric-learning process
as it provides a valuable cue, 
namely, that the three displayed data items are roughly equidistant.


We cast the metric-learning problem as a {\em kernel-learning problem}.
The learned kernel can be used to 
easily compute distances between data items, 
even between data items that did not participate in the metric-learning training phase,
and only their feature vectors are available.
The use of relative comparisons, 
instead of hard ML/CL constraints,
leads to learning a more accurate metric 
that captures relations between data items at different scales.
The learned metric can be used for many data-analysis tasks such as classification, clustering, information retrieval, etc. In this paper, we evaluate our proposed method in a multi-level clustering framework. However, the same method can be applied in many other settings, without loss of generality.

\begin{figure*}[t]
	\begin{center}
	\subfigure[]{\includegraphics[width=0.32\textwidth]{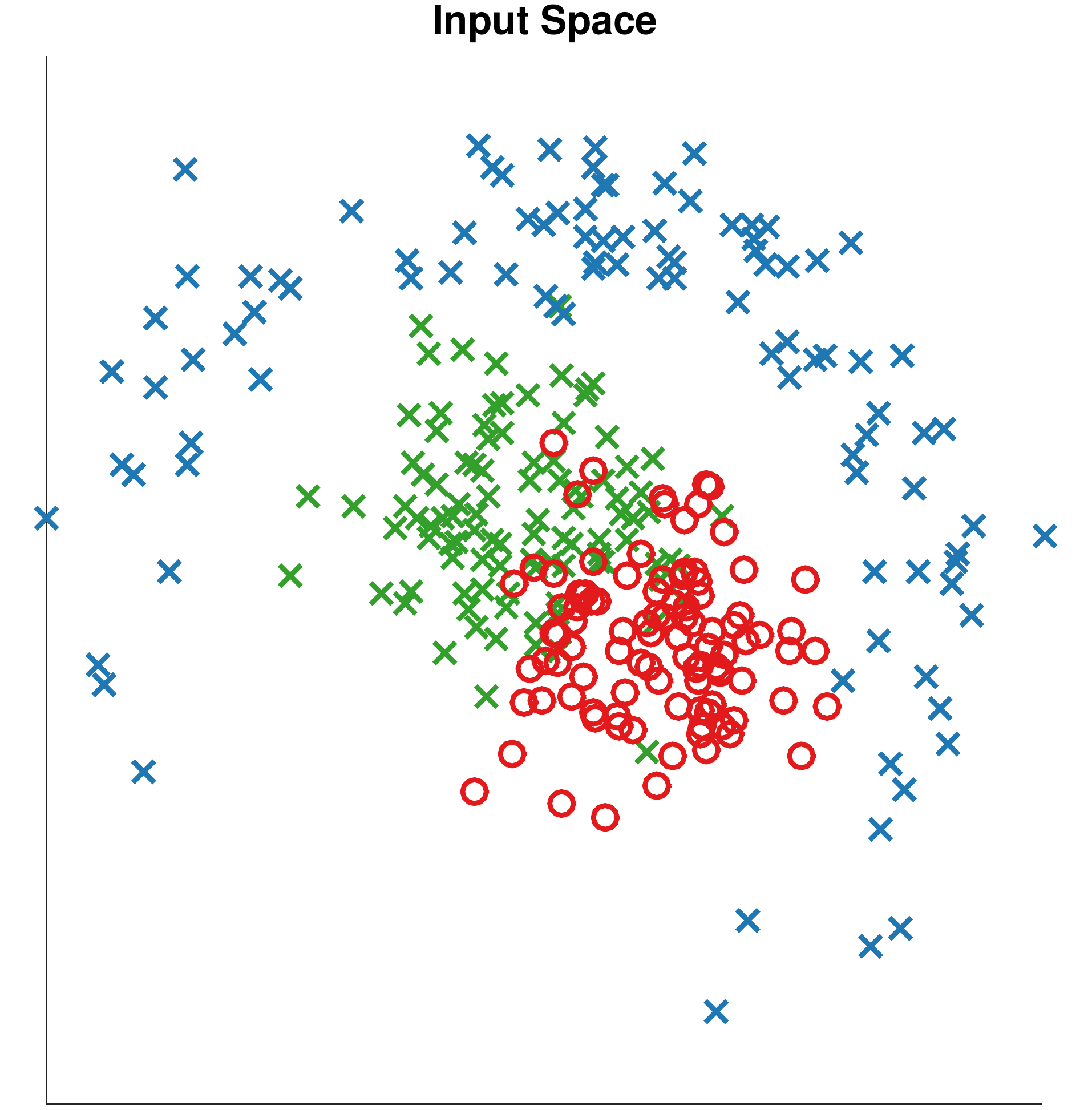}\label{fig:input}}
	\subfigure[]{\includegraphics[width=0.32\textwidth]{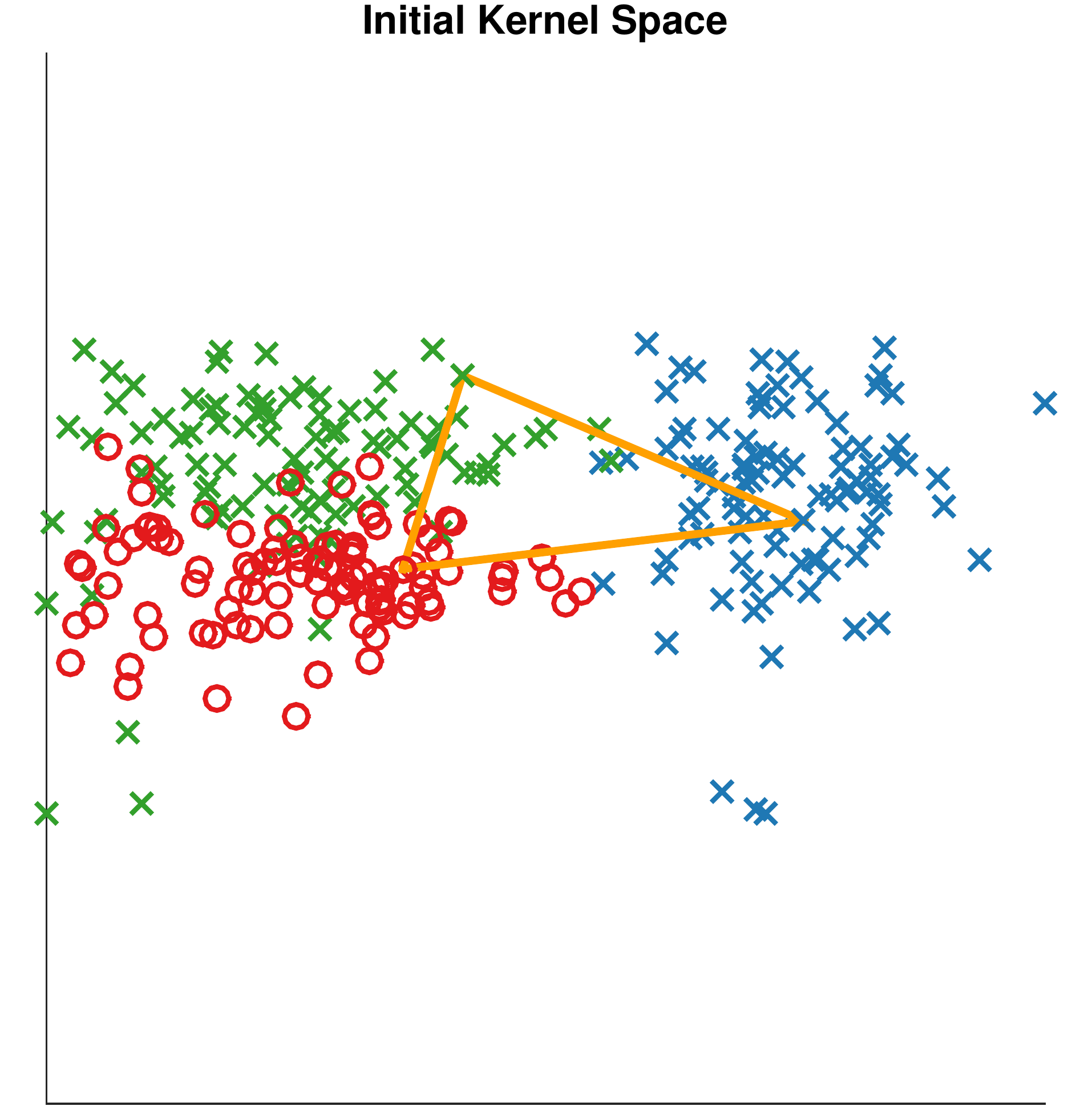}\label{fig:init}}
	\subfigure[]{\includegraphics[width=0.32\textwidth]{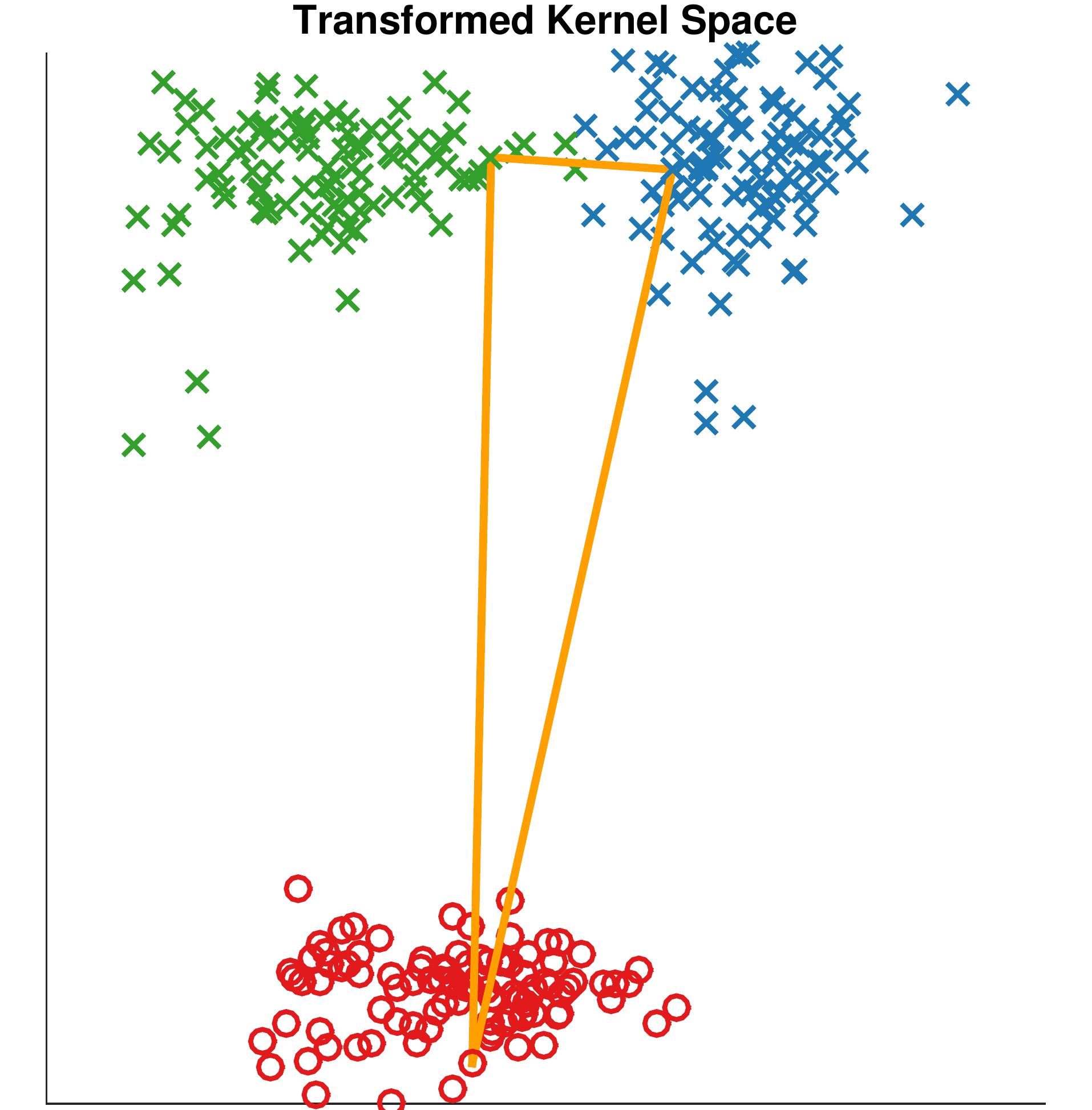}\label{fig:trans}}
	\caption{\label{fig:example} An example of the steps of the algorithm: (a) input space or initial feature representation of a set of items; (b) initial kernel space obtained by applying a non-linear mapping on the input space. The yellow triangle represents a sample triplet, formed by selecting three different items. Each triplet can be selected randomly or in a more systematic manner and the answer to the triplet, which imposes a constraint on the relative-distance of the three items, can be collected from, e.g., a human evaluator. Let us assume that in the shown triplet the red item is an outlier according to the evaluator, i.e., the least similar item to the other two. Clearly this is not satisfied in the initial kernel space. (c) The transformed kernel space in which the triplet in the previous figure is satisfied. In the new representation, the items form clusters in two different resolutions: ($i$) binary class (low resolution): {\sf x}'s vs.\ {\sf o}'s, and 
	($ii$) multiclass (high resolution): each color forms a separate cluster.}
\end{center}
\end{figure*} 

On the technical side, 
we start with an initial kernel $\kernel_0$, 
computed using only the feature vectors of the data items.
We then formulate the kernel-learning task as an optimization problem:
the goal is to find the 
kernel matrix $\kernel$ that is the closest to $\kernel_0$
and satisfies the constraints induced by the relative-comparison labellings.
To solve this optimization task we use known efficient techniques, 
which we adapt for the case of relative comparisons. Figure~\ref{fig:example} illustrates an example of the steps of the algorithm.

More concretely, we make the following contributions:
\begin{enumerate}
\item
We design a kernel-learning method
that can also use unspecified relative distance comparisons.
This is done by extending the method of \citet{skms},
which works with ML and CL constraints.
\item
We introduce a soft formulation that can handle inconsistent constraints. 
The new formulation relies on the use of slack variables. 
To solve the resulting optimization problem 
we develop a new iterative method. 

\item
We perform an extensive experimental validation of our approach
and show that the proposed labeling is indeed more flexible, 
and can lead to a substantial improvement in the clustering accuracy.
We experimentally demonstrate the effectiveness and robustness 
of the new soft-margin formulation in 
handling noisy and inconsistent constraints. 
\end{enumerate}

An earlier version of this work appeared in conference proceedings~\cite{sklr}. 
In this paper, we have extended the previous version of the paper 
by introducing the soft-margin formulation 
and by providing additional experimental results. 

The rest of this paper is organized as follows. 
We start by reviewing the related literature in Section~\ref{sec:related}.
In Section~\ref{sec:prob_form} we introduce our setting and formally define our problem, 
and in Section~\ref{sec:semi_kernel_learning} we present our solution.
In Section~\ref{sec:experiments} we discuss our empirical evaluation, 
and Section~\ref{sec:conclusion} is a short conclusion.

\section{Related Work}
\label{sec:related}

The idea of semi-supervised clustering
was initially introduced by \citet{wagstaff2000}, 
and since then a large number of different problem variants and methods 
have been proposed, the first ones being {\sf COP-Kmeans}
\citep{wagstaff2001} and {\sf CCL} \citep{klein2002}.
Some of the later methods
handle the constraints in a probabilistic framework. 
For instance, 
the ML and CL constraints can be imposed in the form of
a Markov random field prior over the data
items~\citep{basu2004b,basu2004,lu2005}.
Alternatively, \citet{lu2007}
generalizes the standard Gaussian process 
to include the preferences imposed by the ML and CL
constraints.
Recently, \citet{pei2014discriminative}
propose a discriminative clustering model
that uses relative comparisons and,
like our method,
can also make use of unspecified comparisons.

The semi-supervising clustering setting has also been 
studied in the context of spectral clustering, 
and many spectral clustering algorithms have been extended 
to incorporate pairwise constraints~\citep{lu2010,lu2008}. 
More generally, these methods employ techniques 
for semi-supervised graph partitioning and kernel $k$-means algorithms~\citep{dhillon2005}. 
For instance, \citet{kulis2009} present a unified framework for semi-supervised vector and graph clustering using spectral clustering and kernel learning. 

As stated in the introduction,
our work is based on {\em metric learning}.
Most of the metric-learning literature,
starting by the work of \citet{XingNJR02},
aims at finding a Mahalanobis matrix
subject to either ML/CL or relative distance constraints.
\citet{XingNJR02} use ML/CL constraints,
while \citet{SchultzJ03} present a similar approach
to handle relative comparisons. 
Metric learning
often requires solving a semidefinite optimization problem. For instance,~\cite{heim2015relkernel} propose an online kernel learning method using stochastic gradient descent. However, the method requires an additional projection step onto the positive semidefinite cone. 
This problem becomes easier if Bregman divergence,
in particular the log det divergence,
is used to formulate the optimization problem.
Such an approach was first used for metric learning by \citet{davis2007information} with ML/CL constraints,
and subsequently by \citet{LiuMTLL10} likewise with ML/CL constraints,
as well as by \citet{LiuGZJW12} with relative comparisons.
Our algorithm also uses the log det divergence,
and we extend the technique of \citet{davis2007information} 
to handle relative comparisons.

The metric-learning approaches
can also be more directly combined with a clustering algorithm.
The {\sf MPCK-Means} algorithm by \citet{bilenko2004}
is one of the first to combine metric learning with semi-supervised
clustering and ML/CL constraints.
\citet{xiang2008learning}
use metric learning, as well, 
to implement ML/CL constraints
in a clustering and classification framework,
while \citet{KumarK08}
follow a similar approach using relative comparisons.
Recently, \citet{skms}
use a kernel-transformation approach 
to adapt the mean-shift algorithm~\citep{mean_shift} 
to incorporate ML and CL constraints. 
This algorithm, called semi-supervised kernel mean shift clustering ({\sf SKMS}), 
starts with an initial kernel matrix of the data points 
and generates a transformed matrix by minimizing the log det divergence
using an approach based on the work by \citet{lowrank}.
Our paper is largely inspired by the {\sf SKMS} algorithm. 
Our main contribution is to extend the {\sf SKMS} algorithm
so that it handles relative distance comparisons. 

\section{Kernel Learning with Relative Distances}
\label{sec:prob_form}

In this section we
introduce the notation used throughout the paper and formally define the problem we address.

\subsection{Basic Definitions}
Let $\D = \{ 1, \ldots, n \}$ denote a set of {\em data items}
and let $\X = \{\mathbf{x}_i\}_{i=1}^n$, 
with $\mathbf{x}_i \in \mathbb{R}^d$,
denote the set of 
vector representation of these items in a $d$ dimensional Euclidean space;
one vector for every item in~\D.
The vector set~\X\ is the {\em initial feature representation} 
of the items in \D.
We are also given the set~\C\ of {\em relative distance comparisons}
between data items in \D.
These distance comparisons are given in terms of some
{\em unknown distance function}
$\dfun : \D \times \D \rightarrow \mathbb{R}$.
We assume that~\dfun\ reflects certain domain knowledge
that is difficult to quantify precisely,
and can not be directly captured by the features in \X.
Thus, the set of distance comparisons~\C\
\emph{augments our knowledge} about the data items in~\D, 
in addition to the feature vectors in \X.
The comparisons in \C\ are given by human evaluators,
or they may come from some other source.

Given $\X$ and $\C$, our objective is to find a kernel matrix \kernel\
that captures more accurately the distance between data items.
Such a kernel matrix can be used for a number of different purposes. 
In this paper, we focus on using the kernel matrix for clustering the data in~\D.
The kernel matrix \kernel\ is computed by considering both
the similarities between the points in \X\
as well as
the user-supplied constraints induced by the comparisons in \C.

In a nutshell, we compute the kernel matrix \kernel\
by first computing an initial kernel matrix $\kernel_0$ using only the vectors in \X.
The matrix $\kernel_0$ is computed by applying a Gaussian kernel on the vectors in \X.
We then solve an optimization problem in order to 
find the kernel matrix $\kernel$ that is the closest to $\kernel_0$
and satisfies the constraints in \C.

\subsection{Relative-distance Constraints}

The constraints in \C\
express information about distances
between items in \D\
in terms of the distance function \dfun.
However,
we do not need to know the absolute distances between any two items  $i,j\in \D$.
Instead we consider constraints that express information of the type $\dfun(i,j) < \dfun(i,k)$
for some $i,j,k \in \D$.

In particular, every constraint $C \in \C$ is a statement about
the relative distances between {\em three} items in \D.
We consider two types of constraints,
i.e., \C\ can be partitioned into two sets \Crel\ and \Cdk.
The set \Crel\ contains constraints
where one of the three items has been singled out as an ``outlier.''
That is, the distance of the outlying item to the two others
is clearly larger than the distance between the two other items.
The set \Cdk\ contains constraints
where no item appears to be an obvious outlier.
The distances between all three items are then assumed to
be approximately the same.

More formally,
we define \Crel\ to be a set of tuples of the form \trineq{i}{j}{k},
where every tuple is interpreted as 
``item $k$ is an outlier among the three items $i$,~$j$ and~$k$.''
We assume that the item $k$ is an outlier
if its distance from $i$ and~$j$ is at least
\ratio\ times larger than the distance $\dfun(i,j)$, 
for some $\ratio>1$.
This is because we assume
small differences in the distances to be
indistinguishable by the evaluators,
and only such cases end up in \Crel\
where there is no ambiguity between the distances.
Here {\ratio} is a parameter
that must be set in advance by the user.
As a result each triple \trineq{i}{j}{k} in \Crel\ implies the following two inequalities
\begin{eqnarray}
\label{eq:outlier1}
\indneqbig{i}{j}{k}:\quad  \ratio^2\dfun(i,j) &\leq& \dfun(i,k) \;\; \hbox{and}\\
\label{eq:outlier2}
\indneqbig{j}{i}{k}:\quad  \ratio^2\dfun(j,i) &\leq &\dfun(j,k).
\end{eqnarray}
We denote by $\Iset$ the set of all pairs of inequalities imposed by the tuples $\trineq{i}{j}{k} \in \Crel$.

Likewise, 
we define \Cdk\ to be a set of tuples of the form \trieq{i}{j}{k}
that translates to
``the distances between items $i$, $j$ and $k$ are equal.''
In terms of the distance function \dfun,
each triple \trieq{i}{j}{k} in \Cdk\ implies 
\begin{equation}
\label{eq:dontknow}
\dfun(i,j) = \dfun(j,k) = \dfun(i,k).
\end{equation}
Similarly, we denote by $\Eset$ the set of all pairs of equalities imposed by the tuples $\trieq{i}{j}{k} \in \Cdk$.

\subsection{Extension to a Kernel Space}

As mentioned above, 
the first step of our approach is 
forming the initial kernel $\kernel_0$
using the feature vectors \X.
We do this using a standard Gaussian kernel. 
Details are provided in Section~\ref{section:algorithm-wrapup}.

Next we show how the constraints implied by the 
distance comparison sets~\Crel\ and \Cdk\ extend to a kernel space,
obtained by a mapping $\Phi: \D \rightarrow \kspace$.
As usual, we assume that an inner product $\Phi(i)\transpose\Phi(j)$
between items $i$ and $j$ in~\D\ can be
expressed by a symmetric kernel matrix $\kernel$,
that is, $\kernel_{ij} = \Phi(i)\transpose\Phi(j)$.
Moreover,
we assume that the kernel $\kernel$ (and the mapping~$\Phi$)
is connected to the unknown distance function {\dfun}
via the equation
\begin{equation}
\label{eq:dfunkernel}
\dfun(i,j) = \| \Phi(i) - \Phi(j) \|^2 = \kernel_{ii} - 2\kernel_{ij} + \kernel_{jj}.
\end{equation}
In other words,
we explicitly assume that the distance function {\dfun}
is in fact
the Euclidean distance in some unknown vector space.
This is equivalent to assuming that the evaluators 
base their distance-comparison decisions 
on some implicit features,
even if they might not be able to quantify these explicitly.

Next, we discuss the constraint inequalities and equalities
(Equations (\ref{eq:outlier1}), (\ref{eq:outlier2}), and (\ref{eq:dontknow}))
in the kernel space.
Let $\e_i$ denote the vector of all zeros with the value $1$ at position~$i$.
The distance function $\dfun(i,j)$ given in 
Equation~(\ref{eq:dfunkernel}) can be expressed in matrix form
as follows:
\begin{equation*}
\dfun(i,j) =
\kernel_{ii} - 2\kernel_{ij} + \kernel_{jj}
= (\e_i - \e_j)\transpose \kernel (\e_i - \e_j)
= \tr(\kernel (\e_i - \e_j) (\e_i - \e_j)\transpose),
\end{equation*}
where $\tr(\mathbf{A})$ denotes the trace of the matrix $\mathbf{A}$ and we use the fact that $\kernel = \kernel^\top$.
Using the previous equation we can write
Inequality~(\ref{eq:outlier1}) as
\begin{eqnarray*}
\gamma^2 \tr\left(\kernel (\e_i - \e_j) (\e_i - \e_j)\transpose\right) -
   \tr\left(\kernel (\e_i - \e_k) (\e_i - \e_k)\transpose\right) &\leq& 0 \\
\tr\left( \kernel \gamma^2 (\e_i - \e_j) (\e_i - \e_j)\transpose -
  \kernel (\e_i - \e_k) (\e_i - \e_k)\transpose \right) &\leq& 0 \\
\tr\left( \kernel (\gamma^2 (\e_i - \e_j) (\e_i - \e_j)\transpose -
  (\e_i - \e_k) (\e_i - \e_k)\transpose ) \right) &\leq& 0 \\
\tr\left( \kernel\, \Cmat_{\indneq{i}{j}{k}}\right) &\leq& 0,
\end{eqnarray*}
where
$\Cmat_{\indneq{i}{j}{k}} = 
\gamma^2 (\e_i - \e_j) (\e_i - \e_j)\transpose - (\e_i - \e_k) (\e_i - \e_k)\transpose$
is defined to be a matrix that represents the constraint $\indneqbig{i}{j}{k}$.
The matrix $\Cmat_{\indneq{j}{i}{k}}$
defined to represent the constraint $\indneqbig{j}{i}{k}$
for Inequality~(\ref{eq:outlier2})
is formed in exactly the same manner.
Note that unless we set $\ratio > 1$
the Inequalities (\ref{eq:outlier1}) and (\ref{eq:outlier2})
can be satisfied trivially
for a small difference between
the longer and the shorter distance and thus,
the constraint becomes inactive.
Setting $\ratio > 1$ helps avoiding such solutions.

We use a similar technique to
represent the constraints in the set \Cdk.
Recall that the constraint $\trieq{i}{j}{k} \in \Cdk$
implies that the items $i$, $j$, and $k$ are equidistant.
This yields three equations on
the pairwise distances between the items:
$\indeqbig{i}{j}{k}:\dfun(i,j) = \dfun(i,k)$,
$\indeqbig{j}{i}{k}:\dfun(j,i) = \dfun(j,k)$, and
$\indeqbig{k}{i}{j}:\dfun(k,i) = \dfun(k,j)$.
Reasoning as above,
we let
$\Cmat_{\indeq{i}{j}{k}} = 
(\e_i - \e_j) (\e_i - \e_j)\transpose - (\e_i - \e_k) (\e_i - \e_k)\transpose$,
and can thus write the first
equation for the constraint $\trieq{i}{j}{k} \in \Cdk$ as 
\begin{equation*}
\tr(\kernel\, \Cmat_{\indeq{i}{j}{k}}) = 0.
\end{equation*}
The two other equations are written in a similar manner.

\subsection{Log Determinant Divergence for Kernel Learning}
%
Recall that our objective is to find the kernel matrix $\kernel$
that is close to the initial kernel $\kernel_0$.
Assume that $\kernel$ and $\kernel_0$ are both positive semidefinite matrices.
We will use the {\em log determinant divergence}
to compute the similarity between $\kernel_0$ and~$\kernel$.
This is a variant of the {\em Bregman divergence}~\citep{bregman}.

The Bregman divergence between two matrices $\kernel$ and $\kernel_0$ is defined as 
\begin{equation*}
\label{eq:basic_bregman}
\divergence_{\phi}(\kernel,\kernel_0) = \phi(\kernel) -  \phi(\kernel_0) - \tr(\nabla\phi(\kernel_0)^\top (\kernel - \kernel_0)),
\end{equation*}
where $\phi$ is a strictly-convex real-valued function,
and $\nabla\phi(\kernel_0)$ denotes
the gradient evaluated at $\kernel_0$.
Many well-known distance measures are special cases of
the Bregman divergence.
These distance measures can be instantiated by selecting the function $\phi$ appropriately.
For instance, $\phi(\kernel)  = \sum_{ij} K_{ij}^2$
gives the squared Frobenius norm
$\divergence_{\phi}(\kernel,\kernel_0) =  \Vert\kernel- \kernel_0\Vert_F^2$.

For our application in kernel learning,
we are interested in one particular case;
setting $\phi(\kernel) = -\log\det(\kernel)$.
This yields the log determinant (log det) matrix divergence:
\begin{equation}
\label{eq:logdet}
\logdet(\kernel,\kernel_0) = \tr(\kernel\,\kernel_0^{-1}) -  \log\det(\kernel\,\kernel_0^{-1}) - n\, .
\end{equation}
The log det divergence has many interesting properties,
which make it ideal for kernel learning. 
As a general result of Bregman divergences, 
log det divergence is {\em convex} with respect to the first argument. 
Moreover, it can be evaluated using the eigenvalues and eigenvectors of the matrices $\kernel$ and $\kernel_0$. 
This property can be used to extend log det divergence to handle rank-deficient matrices~\citep{lowrank}, 
and we will make use of this in our algorithm described in Section~\ref{sec:semi_kernel_learning}.

\subsection{Problem Definition}
We now have the necessary ingredients to
formulate our semi-supervised kernel learning problem.
Given the set of constraints $\C = \Crel \cup \Cdk$,
the parameter $\ratio$,
and the initial kernel matrix $\kernel_0$,
we aim to find a new kernel matrix {\kernel}, 
which is as close as possible to $\kernel_0$
while satisfying the constraints in \C.
This objective can be formulated as the following constrained minimization problem:
\begin{equation}
\label{eq:minimize}
\begin{aligned}
& \underset{\kernel}{\text{minimize}}\quad \logdet(\kernel,\kernel_0)
& &\\
& \text{subject to} & &\\
& \quad \tr\left( \kernel\, \Cmat_{\indneq{i}{j}{k}}\right) \leq 0, 
\; \tr\left( \kernel\, \Cmat_{\indneq{j}{i}{k}}\right) \leq 0, \quad\mbox{for all }\trineq{i}{j}{k} \in \Crel & \\
& \quad \tr(\kernel\,\Cmat_{\indeq{i}{j}{k}}) = \tr(\kernel\,\Cmat_{\indeq{j}{i}{k}}) = \tr(\kernel\,\Cmat_{\indeq{k}{i}{j}}) = 0,\quad \mbox{for all }\trieq{i}{j}{k} \in \Cdk \\
&\quad \kernel \succeq 0, & & 
\end{aligned}
\end{equation}
where $\kernel \succeq 0$
constrains $\kernel$ to be a positive semidefinite matrix.

\subsection{Relaxation of the Constraints}

The optimization problem~(\ref{eq:minimize}) does not have a solution 
if the set intersection of the inequality and equality constraints 
and the positive semidefinite cone is empty. 
This can happen if there exist a subset of inconsistent relative-comparison constraints. 
In order to be able to handle inconsistent constraints and improve the robustness of our approach,
we formulate a \emph{soft-margin} version of the problem by introducing a set of slack variables $\Xi = \{\xi_{\indneq{i}{j}{k}}, \xi_{\indeq{i}{j}{k}}\}$. 
In more detail, the soft-margin problem formulation asks to
\begin{equation}
\label{eq:minimize_soft}
\begin{aligned}
& \underset{\kernel,\, \Xi}{\text{minimize}}\quad \logdet(\kernel,\kernel_0) + \frac{1}{2}\,\lambda_{\text{neq}} \sum_{ \indneq{i}{j}{k} \in \Iset} \xi^2_{\indneq{i}{j}{k}} +  \frac{1}{2}\,\lambda_{\text{eq}} \sum_{ \indeq{i}{j}{k} \in \Eset} \xi^2_{\indeq{i}{j}{k}}, 
& &\\
& \text{subject to} & &\\
& \quad \tr\left( \kernel\, \Cmat_{\indneq{i}{j}{k}}\right) \leq \xi_{\indneq{i}{j}{k}}, 
\; \tr\left( \kernel\, \Cmat_{\indneq{j}{i}{k}}\right) \leq \xi_{\indneq{j}{i}{k}}, 
\;\mbox{for all }\trineq{i}{j}{k} \in \Crel& & \\
& \quad \tr(\kernel\,\Cmat_{\indeq{i}{j}{k}}) = \xi_{\indeq{i}{j}{k}}, 
\; \tr(\kernel\,\Cmat_{\indeq{j}{i}{k}}) = \xi_{\indeq{j}{i}{k}},\; 
\tr(\kernel\,\Cmat_{\indeq{k}{i}{j}}) = \xi_{\indeq{k}{i}{j}},\; \mbox{for all }\trieq{i}{j}{k} \in \Cdk & &\\
&\quad \kernel \succeq 0, & & 
\end{aligned}
\end{equation}
in which, $\xi_{\indneq{i}{j}{k}}$ and $\xi_{\indeq{i}{j}{k}}$ are the slack variables 
associated with the inequality constraint $\indneqbig{i}{j}{k}$ 
and the equality constraint $\indeqbig{i}{j}{k}$, respectively. 
Note that the inequality slack variables $\xi_{\indneq{i}{j}{k}}$ must be 
non-negative,\footnote{However, we do not need to impose this constraint implicitly 
since a solution with $\xi_{\indneq{i}{j}{k}}  < 0$ cannot be optimum.}
while no such non-negativity condition is required for the equality slack variables $\xi_{\indeq{i}{j}{k}}$.
Finally, $\lambda_{\text{eq}}$ and $\lambda_{\text{neq}}$ are regularization parameters 
that control the trade-off between minimizing the divergence
and minimizing the magnitude of the slack variables.

\section{Semi-supervised Kernel Learning}
\label{sec:semi_kernel_learning}

We now focus on the optimization problems defined above, Problems~(\ref{eq:minimize}) and (\ref{eq:minimize_soft}).
It can be shown that in order to have a finite value for the log det divergence,
the rank of the matrices must remain equal~\citep{lowrank}. 
This property along with the fact 
that the domain of the log det divergence is the positive-semidefinite matrices, 
allow us to perform the optimization 
without explicitly restraining the solution to the positive-semidefinite cone 
nor checking for the rank of the solution. 
This is in contrast
with performing the optimization using, say, the Frobenius norm,
where the projection to the positive semidefinite cone must be explicitly imposed.

\subsection{Bregman Projections for Constrained Optimization}
\label{subsec:bregman_proj}

In solving the optimization Problem~(\ref{eq:minimize}), 
the aim is to minimize the divergence 
while satisfying the set of constraints imposed by $\C= \Crel \cup \Cdk$. 
In other words, we seek for a kernel matrix $\kernel$ 
by projecting the initial kernel matrix $\kernel_0$ 
onto the convex set obtained from the intersection of the set of constraints. 
The optimization Problem~(\ref{eq:minimize}) 
can be solved using the method of \emph{Bregman projections}~\citep{lowrank,bregman,tsuda2005}. 
The idea is to consider one unsatisfied constraint at a time 
and project the matrix so that the constraint gets satisfied. 
Note that the projections are not orthogonal and thus, 
a previously satisfied constraint might become unsatisfied. 
However, as stated before, the objective function in Problem~(\ref{eq:minimize}) 
is convex and the method is guaranteed to converge to the global minimum 
if all the constraints are met infinitely often (randomly or following a more structured procedure).

Let us consider the update rule for an unsatisfied constraint from \Crel. 
We first consider the case of full-rank symmetric positive semidefinite matrices. 
Let $\kernel_t$ be the value of the kernel matrix at step~$t$. 
For an unsatisfied inequality constraint 
$\Cmat$,
the optimization problem becomes\footnote{We skip the subscript for notational simplicity.}
\begin{equation}
\label{eq:sub_minimize}
\begin{aligned}
& \kernel_{t+1} = \underset{\kernel}{\text{arg\,min }} \logdet(\kernel,\kernel_t), \\
\text{subject to}\;\;
& \langle \kernel,\Cmat \rangle = \tr(\kernel \Cmat) \leq 0.
\end{aligned}
\end{equation}
Using a Lagrange multiplier $\alpha \geq 0$, we can write
\begin{equation}
\label{eq:lagrange_form}
 \kernel_{t+1} = 
 \underset{\kernel}{\text{arg\,min }} \logdet(\kernel,\kernel_t) + \alpha \tr(\kernel \Cmat).
 \end{equation}
Taking the derivative of Equation~\eqref{eq:lagrange_form} with respect to $\kernel$ and setting it to zero, we have the update

\begin{equation}
\label{eq:kernel_opt}
\kernel_{t+1}  = (\kernel_{t}^{-1} + \alpha \Cmat)^{-1}.
\end{equation}

In order to determine the value of $\alpha$, we substitute the update Equation~\eqref{eq:kernel_opt} into~\eqref{eq:lagrange_form} and form the conjugate dual optimization problem with respect to $\alpha$. This simplifies to the following optimization problem (see Appendix~A)
\begin{equation}
\label{eq:alpha_const}
\begin{aligned}
& \alpha^* = \arg\max_{\alpha} \log\det(\mathbf{I}_n + \alpha\,\kernel_t\,\Cmat), \\
\text{subject to}\;\;
& \alpha \geq 0.
\end{aligned}
\end{equation}
in which $\mathbf{I}_n$ is the $n\times n$ identity matrix.

Equation (\ref{eq:alpha_const}) does not have a closed form solution for $\alpha$, in general.
However, we exploit the fact that both types of our constraints, 
the matrix \Cmat\ has rank 2, i.e., $\text{rank}(\Cmat) = 2$. 
Let $\eta_1$ and $\eta_2$ be the eigenvalues of the matrix product $\kernel_t \Cmat$.
It can be shown that $\eta_2\leq 0 \leq \eta_1$ and $|\eta_2| \leq |\eta_1|$. Thus, Equation~\eqref{eq:alpha_const} can be written as follows
\begin{equation}
\label{eq:alpha_const_eig_form}
\begin{aligned}
& \alpha^* = \arg\max_{\alpha}\, \log(1+\alpha\eta_1)(1+\alpha\eta_2), \\
\text{subject to}\;\;
& \alpha \geq 0.
\end{aligned}
\end{equation}
Solving Equation~\eqref{eq:alpha_const_eig_form} for $\alpha^*$ gives 
\begin{equation*}
\label{eq:alpha_eig}
\frac{\eta_1}{1+\alpha^*\eta_1} +  \frac{\eta_2}{1+\alpha^*\eta_2} = 0 ,
\end{equation*}
and
\begin{equation}
\label{eq:alpha_sol}
\alpha^* = -\frac{1}{2}\frac{\eta_1 + \eta_2}{\eta_1\eta_2}\geq 0 .
\end{equation}
Finally, substituting for Equation (\ref{eq:alpha_sol}) for $\alpha$ into (\ref{eq:kernel_opt}), 
yields the update equation for the kernel matrix.
The update Equation~\eqref{eq:kernel_opt} can be simplified further using some matrix algebra.
Let $\Cmat = \mathbf{UV}^\top$ where $\mathbf{U,V}$ are $n\times 2$ matrices of rank-$2$
\begin{equation*}
\mathbf{U} = [\gamma(\mathbf{e}_i - \mathbf{e}_j),\, (\mathbf{e}_i - \mathbf{e}_k)],\quad \mathbf{V} = [\gamma(\mathbf{e}_i - \mathbf{e}_j),\, -(\mathbf{e}_i - \mathbf{e}_k)]
\end{equation*}

Let $\kernel_t = \mathbf{G}_t\,\mathbf{G}_t^\top$ be the Cholesky decomposition of the kernel matrix. Using Sherman-Morrison-Woodbury formula~(\cite{sherman,golub}), we can write~(\ref{eq:kernel_opt}) as
\begin{equation}
\label{eq:kernel_update_modified}
\begin{split}
\kernel_{t+1}   &= \kernel_t - \alpha^*\,\kernel_t\, \mathbf{U}\, (\mathbf{I} + \alpha^* \mathbf{V}^\top\, \kernel_t\,  \mathbf{U} )^{-1}\, \mathbf{V}^\top\, \kernel_t\\
&= \mathbf{G}_t\, \left(\mathbf{I}_n - \alpha^*\,\mathbf{G}_t^\top\, \mathbf{U}\, (\mathbf{I} + \alpha^* \mathbf{V}^\top\, \kernel_t\,  \mathbf{U} )^{-1}\, \mathbf{V}^\top\, \mathbf{G}_t\right)\, \mathbf{G}_t^\top\\
& = \mathbf{G}_t\, \mathbf{W}\,\mathbf{W}^\top\, \mathbf{G}_t^\top\\
& = \mathbf{G}_{t+1} \mathbf{G}_{t+1}^\top
\end{split}
\end{equation}
where $\mathbf{W}\,\mathbf{W}^\top$ is the Cholesky decomposition of the diagonal plus rank-2 update term which can be efficiently calculated in $\mathcal{O}(n)$ time (see Appendix~C). Thus, the computational bottleneck is the calculation of the matrix product $\mathbf{G}_{t+1} = \mathbf{G}_t\, \mathbf{W}$.

For an equality constraint \Cdk, $\alpha^*$ must satisfy the following constraint
\begin{equation}
\label{eq:alpha_eq}
\tr((\kernel_t^{-1} + \alpha^* \Cmat)^{-1} \Cmat) = 0.
\end{equation}
The value of $\alpha^*$ satisfying Equation~\eqref{eq:alpha_eq} can be written as the stationary point of the following function
\begin{equation*}
-\log\det(\kernel_t^{-1} + \alpha\, \Cmat) = -\log\det(\mathbf{I}_n + \alpha\, \kernel_t\,\Cmat) + \text{const.}
\end{equation*}
which is the same as stationary point of Equation~\eqref{eq:alpha_const}. Thus, $\alpha^*$ for an equality constraint is calculated similar to a inequality constraint using Equation~\eqref{eq:alpha_sol}. In oder words, the Bregman projection for an inequality constraint projects the kernel matrix onto the boundary of the convex set of matrices that satisfy the given constraint.

For a rank-deficient kernel matrix 
$\kernel_0$ with $\text{rank}(\kernel_0)=r$, 
we employ the results of \citet{lowrank},
which state that for any column-orthogonal matrix $\mathbf{Q}$ 
with $\text{range}(\kernel_0) \subseteq \text{range}(\mathbf{Q})$ 
(e.g., obtained by singular value decomposition of $\kernel_0$),
we first apply the transformation 
\[
\mathbf{M} \rightarrow \mathbf{\hat{M}} = \mathbf{Q}^\top\, \mathbf{M\, Q},
\] 
on all the matrices, 
and after finding the kernel matrix 
$\mathbf{\hat{K}}$ satisfying all the transformed constraints, 
we can obtain the final kernel matrix using the inverse transformation 
\[
\kernel = \mathbf{Q\,\hat{K}\, Q^\top}.
\]
Note that since log det preserves the matrix rank, 
the mapping is one-to-one and invertible.
 
\subsection{Bregman Projections for Soft Margin Formulation}
\label{subsec:bregman_proj_soft}
 
We now consider solving the soft-margin formulation defined as Problem~(\ref{eq:minimize_soft}). 
Similar to the previous case, 
the method of Bregman projections can be applied iteratively 
by considering one unsatisfied constraint at a time and performing the update. 
In our exposition below, we consider the update rule for an inequality constraint. 
The update rule for an equality constraint applies in a similar manner.

Let us denote by $\kernel_t$ the value of the kernel matrix at step $t$. 
Considering an unsatisfied inequality constraint $\Crel$, now the optimization becomes
\begin{equation}
\label{eq:sub_minimize_soft}
\begin{aligned}
& \underset{\kernel,\, \xi}{\text{minimize}}
&&\logdet(\kernel,\kernel_t) + \frac{1}{2}\, \lambda\, \xi^2, \\
& \text{subject to}\;\;
&& \langle \kernel,\Cmat \rangle = \tr(\kernel^\top \Cmat) \leq \xi, \\
\end{aligned}
\end{equation} 
where $\xi$ is the non-negative slack variable associated with $\Cmat$ and $\lambda$ is the regularization factor.
Using the Lagrange multiplier method, 
the optimization Problem~(\ref{eq:sub_minimize_soft}) can be written so as the expression
\begin{equation}
\label{eq:lagrange_form_soft}
 \logdet(\kernel,\kernel_t) + \frac{1}{2}\,\lambda\, \xi^2 + \alpha\, \left(\tr(\kernel^\top \Cmat)-\xi\right)
 \end{equation} 
is minimized with respect to $\kernel$ and $\xi$, and maximized with respect to $\alpha$.
Lagrange multiplier $\alpha$ must also satisfy the set of Karush-Kuhn-Tucker (KKT) conditions
\begin{eqnarray*}
\alpha\, \left(\tr(\kernel^\top \Cmat)-\xi \right) &=& 0\label{eq:alpha_kkt1}\\
\alpha &\geq & 0\,.\label{eq:alpha_kkt2}
\end{eqnarray*}
Setting the derivative of the expression~(\ref{eq:lagrange_form_soft}) equal to zero, 
with respect to $\kernel$ and $\xi$, yields the following two equations
\begin{eqnarray}
\kernel_{t+1}  &=& (\kernel_{t}^{-1} + \alpha \Cmat)^{-1}\label{eq:kernel_opt_soft} \;\mbox{ and}\\
\lambda\,\xi &=& \alpha \label{eq:xi_alpha_mu}\,.
\end{eqnarray}
Substituting for $\kernel_{t+1}$ and $\xi$ in Equation~\eqref{eq:lagrange_form_soft} using Equations~\eqref{eq:kernel_opt_soft} and~\eqref{eq:xi_alpha_mu}, respectively, we have the following dual optimization problem over $\alpha$ (see Appendix B)
\begin{equation}
\label{eq:alpha_dual_soft}
\begin{aligned}
& \alpha^* = \arg\max_{\alpha} \log\det(\mathbf{I}_n + \alpha\,\kernel_t\,\Cmat) - \frac{1}{2} \lambda\, \alpha^2, \\
\text{subject to}\;\;
& \alpha \geq 0.
\end{aligned}
\end{equation}

Equation~(\ref{eq:alpha_dual_soft}) can be solved similar to Equation~(\ref{eq:alpha_const_eig_form}) 
using the eigenvalues of the product matrix $\kernel_t\, \Cmat$, 
that is
\begin{equation}
\label{eq:alpha_eig_soft}
\frac{\eta_1}{1+\alpha\eta_1} +  \frac{\eta_2}{1+\alpha\eta_2} - \lambda\alpha = 0
\end{equation}
Equation~(\ref{eq:alpha_eig_soft}) yields to a cubic polynomial equation
\begin{equation}
\label{eq:poly}
\alpha^3 + \frac{\eta_1+\eta_2}{\eta_1\eta_2}\, \alpha^2 + \left(\frac{1}{\eta_1\eta_2}-2\lambda\right)\, \alpha - \frac{\eta_1\eta_2}{\eta_1+\eta_2} = 0\, .
\end{equation}
The polynomial Equation~(\ref{eq:poly}) has two positive roots and one negative root since the sum of the roots 
(the coefficient of the quadratic term) is negative and the multiplication of the roots (the constant term) is positive. To choose the appropriate positive root of the polynomial, 
we note that Equation~(\ref{eq:alpha_eig_soft}) is consistent with Equation~(\ref{eq:alpha_eig}) when the constraints are forced to be exactly satisfied, i.e., when the penalty $\lambda$ for an unsatisfied constraint is very large. Thus, when $\frac{\alpha}{\lambda} \ll 1$ holds for large values of $\lambda$, the solution of Equation~(\ref{eq:alpha_eig_soft}) can be seen as a perturbation on the value of $\alpha^*$, 
obtained using Equation~(\ref{eq:alpha_sol}). 
Using the first-order expansion near $\alpha^*$, 
we can write the solution of Equation~(\ref{eq:alpha_eig_soft}) as 
\begin{equation}
\label{eq:alpha_sol_pert}
\hat{\alpha}^* = \alpha^* - \delta\alpha + \mathcal{O}(({\scriptstyle \frac{\alpha^*}{\lambda}})^2),\;\; \delta\alpha = \frac{1}{8}\frac{(\eta_1-\eta_2)^2}{(\eta_1\eta_2)^2}\, \frac{\alpha^*}{\lambda} \geq 0.
\end{equation}
Note that from Equation~(\ref{eq:alpha_eig_soft}), for large values of $\vert\alpha\eta_i \vert\gg 1,\, i = 1, 2$, the roots of the polynomial approach the value $\pm\sqrt{2\lambda}$. 
Thus, the smaller positive root of the polynomial corresponds the desired $\alpha^*$ for the update. The update for an equality constraint follows similarly, as in the previous case.

After obtaining the proper value of $\alpha^*$ from solving Equation~(\ref{eq:poly}), 
the update rule for the kernel matrix can be applied similarly, using Equation~(\ref{eq:kernel_update_modified}). We repeat the procedure until the values of $\alpha$ for all constraints stabilize 
(the values of $\xi$ converge).

\subsection{Relations to Metric Learning and Generalization to Out-of-sample Data }

In semi-supervised kernel-learning, the set of constrained data points may be significantly smaller than the set of all data points. Thus, applying the kernel updates on the full kernel matrix may become computationally expensive. Additionally, the learned kernel function, satisfying the set of constraints, may need to be generalized to a set of new (unseen) data points. Our kernel-learning method can handle both cases by bridging the kernel-learning problem with an equivalent metric-learning problem in a transformed kernel space.

\citet{jain} demonstrated the equivalence between the kernel-learning problem (Problem~(\ref{eq:minimize})) 
and the following metric-learning problem using a linear transformation in the feature space corresponding to the mapping 
$\Phi_0: \D \rightarrow \kspace$ of the initial kernel space

\begin{equation}
\label{eq:metric_learn}
\begin{aligned}
& \underset{\W}{\text{minimize}}\quad \logdet(\W,\mathbf{I})
& &\\
& \text{subject to} & &\\
& \quad \tr( \W\, \Cmat'_{\indneq{i}{j}{k}}) \leq 0, 
\; \tr( \W\, \Cmat'_{\indneq{j}{i}{k}}) \leq 0, \;\mbox{ for all }\trineq{i}{j}{k} \in \Crel& & \\
& \quad \tr(\W\,\Cmat'_{\indeq{i}{j}{k}}) = 0, 
\; \tr(\W\,\Cmat'_{\indeq{j}{i}{k}}) = 0,\; \tr(\W\,\Cmat'_{\indeq{k}{i}{j}}) = 0,\;\mbox{ for all }\trieq{i}{j}{k} \in \Cdk \\
&\quad \W \succeq 0, & & 
\end{aligned}
\end{equation}
where $\Cmat'_{\indneq{i}{j}{k}}$ is defined as
\begin{equation}
\label{eq:metric_learn_const}
\Cmat'_{\indneq{i}{j}{k}} = \gamma \left(\Phi_0(i) - \Phi_0(j)\right)\left(\Phi_0(i) - \Phi_0(j)\right)^\top - \left(\Phi_0(i) - \Phi_0(k)\right)\left(\Phi_0(i) - \Phi_0(k)\right)^\top .
\end{equation}
The soft-margin formulation using the slack variables in Problem~(\ref{eq:minimize_soft}) 
can be written in a similar manner. 
Note that the Problem~(\ref{eq:metric_learn}) is defined in the feature space imposed by $\Phi_0$,
and thus, 
the matrix $\W$ can be of infinite dimension. 
However, the problem can be solved implicitly by solving the kernel-learning Problem~(\ref{eq:minimize}),
in which $\kernel_0 = \boldsymbol{\Phi}_0^\top \boldsymbol{\Phi}_0$, 
and where $\boldsymbol{\Phi}_0$ is the feature matrix 
whose $i$-th column is equal to $\Phi_0(i)$ and the optimal solutions are related by the following 
equations
\begin{align}
\kernel &= \boldsymbol{\Phi}_0^\top \W \boldsymbol{\Phi}_0 ,\label{eq:kernel_w}\\
\W &= \mathbf{I} + \boldsymbol{\Phi}_0 \kernel_0^{-1} (\kernel - \kernel_0) \kernel_0^{-1} \boldsymbol{\Phi}_0 .\label{eq:w_kernel}
\end{align}
Note that for the identity mapping $\Phi_0(i) = \mathbf{x}_i$, 
Problem (\ref{eq:metric_learn}) reduces to a linear transformation of the vector in the original space.

As a result,
the kernel matrix learned by minimizing the log det divergence 
subject to the set of constraints  $\Crel \cup \Cdk$ 
can be also extended to handle \emph{out-of-sample} data points, 
i.e., data points that were not present when learning the kernel matrix. 
Using~\eqref{eq:kernel_w} and~\eqref{eq:w_kernel}, the inner product between a pair of out-of-sample data points
$\mathbf{x}, \mathbf{y} \in \mathbb{R}^d$ in the transformed kernel space can written as
\begin{equation}
\label{eq:out_of_sample}
k(\mathbf{x},\mathbf{y}) 
= k_0(\mathbf{x},\mathbf{y}) 
+ \mathbf{k_x}^\top (\kernel_0^\dagger\, (\kernel-\kernel_0)\,\kernel_0^\dagger)\,\mathbf{k_y} ,
\end{equation}
where the value of $k_0(\mathbf{x},\mathbf{y})$ and the vectors $\mathbf{k_x} = [k_0(\mathbf{x},\mathbf{x}_1), \ldots, k_0(\mathbf{x},\mathbf{x}_n)]^\top$ and $\mathbf{k_y} = [k_0(\mathbf{y},\mathbf{x}_1), \ldots, k_0(\mathbf{y},\mathbf{x}_n)]^\top$ are computed using the initial kernel function. 
This observation allows learning the kernel matrix by using only the subset of constrained data points and thus, avoiding the computational overhead in large datasets.

\subsection{Semi-supervised Kernel Learning with Relative Comparisons}
\label{section:algorithm-wrapup}
 
In this section, 
we summarize the proposed approach, which
we name {\sf SKLR}, 
for  
Semi-supervised Kernel-Learning with Relative comparisons. We refer to the soft formulation of the algorithm as {\sf soft SKLR} (or {\sf sSKLR}, for short).
The pseudo-code of the {\sf SKLR} method is shown in Algorithm~\ref{alg:main}.\footnote{Note that the algorithms for {\sf SKLR} and {\sf sSKLR} essentially differ only on the value of $\alpha$ for update.}
As already discussed, the main ingredients of the method are the following.
 
\subsubsection{Selecting the Bandwidth Parameter for $k_0$.}

We consider an adaptive approach to select the bandwidth parameter of the Gaussian kernel function. 
First, we set $\sigma_i$ equal to the distance between point $\mathbf{x}_i$ 
and its $k$-th nearest neighbor. 
Next, we set the kernel between $\mathbf{x}_i$ and $\mathbf{x}_j$ to
\begin{equation}
\label{eq:adaptive_kernel}
k_0(\mathbf{x}_i,\mathbf{x}_j) 
= \exp \left( -\frac{\Vert\mathbf{x}_i-\mathbf{x}_j\Vert^2}{\sigma_{ij}^2}\right),
\end{equation}
where, $\sigma_{ij}^2 = \sigma_{i}\sigma_{i}$. 
This process ensures a large bandwidth for sparse regions and a small bandwidth for dense regions.

\subsubsection{Semi-supervised Kernel learning with Relative Comparisons.}

After finding the low-rank approximation of the initial kernel matrix $\kernel_0$ 
and transforming all the matrices by a proper matrix $\mathbf{Q}$, 
as discussed in Section~\ref{subsec:bregman_proj}, 
the algorithm proceeds by randomly considering one unsatisfied constrained at a time  
and performing the Bregman projections~(\ref{eq:kernel_update_modified}) 
until all the constraints are satisfied (or alternatively, the slack variables converge).
 
\subsubsection{Clustering Method.}

After obtaining the kernel matrix $\kernel$ 
satisfying the set of all relative and undetermined constraints, 
we can obtain the final clustering of the points by applying any standard kernelized clustering method. 
In this paper, we consider the kernel $k$-means because of its simplicity and good performance. 
Generalization of the method to other clustering techniques such as kernel mean-shift is straightforward.

\subsubsection{Computational Complexity}

The computation complexity of the algorithm is dominated by the Bregman projection step, given in Equation~(\ref{eq:kernel_update_modified}). This step requires $\mathcal{O}(n^2)$ flops since both $\mathbf{U}$ and $\mathbf{V}$ are $n \times 2$ matrices. On the other hand, working with the rank-deficient matrices, we can reduce the complexity to $\mathcal{O}(r^2)$ where $r \leq n$. It is also possible to generalize the approach in~(\cite{lowrank}) to work on a factorized form of the update equation~(\ref{eq:kernel_update_modified}) by factoring the kernel matrix $\kernel$ into $\mathbf{GG}^\top$. However, due to the cross terms in rank-$2 $ updates, it is not possible to directly obtain the right multiplication of the update term and the low-rank factor (see \citet{lowrank}, Algorithm 1). As the final remark, we note that it is possible to consider only the subset of the constrained data points in the kernel learning step and then, generalize the learned kernel to the out-of-sample data points using Equation~(\ref{eq:out_of_sample}). This can significantly reduce the size of the kernel learning problem and thus, result in a major speed-up.
 
 \begin{algorithm}[t]
   \caption{   \label{alg:main} ({\sf SKLR}) 
   Semi-supervised kernel learning with relative comparisons}

\begin{algorithmic}
   \STATE {\bfseries Input:} initial $n\times n$ kernel matrix $\kernel_0$, 
   set of relative comparisons $\Crel$ and $\Cdk$, constant distance factor $\gamma$
   \STATE {\bfseries Output:} kernel matrix $\kernel$
   \bigskip
   
   \STATE $\bullet$ \textbf{Find low-rank representation:}
   \begin{itemize}
   \setlength\itemsep{0.1mm}
   \item Compute the $n\times n$ low-rank kernel matrix $\mathbf{\tilde{K}}_0$ such that $\text{rank}(\mathbf{\tilde{K}}_0) = r \leq n$ using singular value decomposition such that $\frac{\Vert\mathbf{\tilde{K}}_0\Vert_F}{\Vert\kernel_0\Vert_F} \geq 0.9$\\
   \item Set $\kernel_0  \gets \mathbf{\tilde{K}}_0$\\
   \item Find $n\times r$ column orthogonal matrix $\mathbf{Q}$ such that $\text{range}(\kernel_0) \subseteq \text{range}(\mathbf{Q})$\\
   \item Apply the transformation $\mathbf{\hat{M}} \gets \mathbf{Q}^\top\,\mathbf{M\,Q}$ on all matrices
   \end{itemize}
   
   \STATE $\bullet$ \textbf{Initialize the kernel matrix} 
   \begin{itemize}
   \setlength\itemsep{0.1mm}
   \item Set $\mathbf{\hat{K}} \gets \mathbf{\hat{K}}_0$
    \end{itemize}
   
   $\bullet$ \textbf{Repeat}
   \begin{itemize}
   \setlength\itemsep{0.1mm}
   \item[] (1) Select an unsatisfied constraint $\mathbf{\hat{C}} \in \Crel \cup \Cdk$\\
   \item[] (2) Apply Bregman projection (\ref{eq:kernel_update_modified})\\
   \end{itemize}
   \STATE \hspace{3mm}\textbf{Until} all the constraints are satisfied 
   \STATE $\bullet$ \textbf{Return} $\kernel \gets \mathbf{Q\, \hat{K}\, Q^\top}$
\end{algorithmic}
\end{algorithm}

\section{Experimental Results}
\label{sec:experiments}

In this section, we evaluate the performance of the proposed kernel-learning method,
{\sf SKLR} along with its soft margin counterpart, {\sf sSKLR}.
As the under-the-hood clustering method required by {\sf SKLR}, 
we use the standard kernel $k$-means with Gaussian kernel and without any supervision 
(Equation (\ref{eq:adaptive_kernel})).
We compare {\sf SKLR} and {\sf sSKLR} to three different semi-supervised metric-learning algorithms, 
namely, 
{\sf ITML}~\citep{davis2007information},
{\sf SK$k$m}~\citep{skms} (a variant of {\sf SKMS} with kernel $k$-means in the final stage), 
and {\sf LSML}~\citep{LiuGZJW12}. 
We select the {\sf SK$k$m} variant as
\citet{skms} have shown that {\sf SK$k$m} 
tends to produce more accurate results than other semi-supervised clustering methods.
Two of the baselines, 
{\sf ITML} and {\sf SK$k$m},
are based on pairwise ML/CL constraints, 
while {\sf LSML} uses relative comparisons. 
For {\sf ITML} and {\sf LSML}
we apply $k$-means on the transformed feature vectors to find the final clustering, 
while for {\sf SK$k$m}, {\sf SKLR}, and {\sf sSKLR}, we apply kernel $k$-means on the transformed kernel matrices. 

To assess the quality of the resulting clusterings, 
we use the Adjusted Rand (AR) index~\citep{AR}. 
Each experiment is repeated $30$ times and the average over all trials is reported. 
For the parameter {\ratio} required by {\sf SKLR} and {\sf sSKLR} we use $\ratio^2 = 2$. We also use $\lambda_{\text{neq}} = \lambda_{\text{eq}} = 10^5$ for the regularization parameters required by the {\sf sSKLR} method\footnote{We did not observe significant difference in the performance when varying $\lambda$ is the range $[10^3,10^5]$.}.
In the following,
all experiments are conducted with noise-free constraints,
with the exception of the one discussed in Section~\ref{sect:noise}.
Our implementation of {\sf SKLR} and {\sf sSKLR} is in MATLAB and the code is publicly 
available.\footnote{\url{https://github.com/eamid/sklr}} 
For the other three methods we use publicly available implementations.\footnote{\url{http://www.cs.utexas.edu/~pjain/itml}}\footnote{\url{https://github.com/all-umass/metric_learn}}\footnote{\url{https://www.iiitd.edu.in/~anands/files/code/skms.zip}}

\subsection{Datasets}

We conduct the experiments on three different real-world datasets.\smallskip\\ 
\textbf{Vehicle:\footnote{\url{http://www.csie.ntu.edu.tw/~cjlin/libsvmtools/datasets/}}} 
The dataset contains $846$ instances from $4$ different classes 
and is available on the LIBSVM repository.\smallskip\\
\textbf{MIT Scene:\footnote{\url{http://people.csail.mit.edu/torralba/code/spatialenvelope/}}} 
The dataset contains $2688$ outdoor images, each sized $256\times 256$, 
from $8$ different categories: 
$4$ natural and $4$ man-made. 
We use the GIST descriptors~\citep{gist} as the feature vectors.\smallskip\\
\textbf{USPS Digits:\footnote{\url{http://cs.nyu.edu/~roweis/data.html}}} 
The dataset contains $16\times 16$ grayscale images of handwritten digits. 
It contains $1100$ instances from each class. 
The columns of each images are concatenated to form a $256$ dimensional feature vector.

\begin{table}[!t]
\centering
\begin{tabular}{lcccccc}
\toprule
Dataset & {\sf K$k$m} & {\sf LSML} & {\sf ITML} & {\sf SK$k$m} & {\sf SKLR} & {\sf sSKLR}\\
\midrule
\footnotesize{{\bf Vehicle} (binary)} & -0.0504  &  0.0748  &  0.0938  &  0.2312  &  {\bf 0.4706}  &  0.3633\\
\footnotesize{{\bf MIT Scene} (binary)} & 0.0099  &  0.4466   & 0.4792  &  0.5502  &  {\bf 0.8198}  &  0.7998\\ 
\footnotesize{{\bf USPS Digits} (binary)} & 0.0194  &  0.2864 &   0.3385  &  0.4049  &  {\bf 0.7147} &   0.7054\\
\footnotesize{{\bf Vehicle} (multiclass)} & 0.0541 &   0.0789   & 0.0780  &  0.1338  &  {\bf 0.2582}  &  0.2167
\\
\footnotesize{{\bf MIT Scene} (multiclass)} & 0.3363  &  0.4494  &  0.3681  &  0.4062  &  0.4609  &  {\bf 0.5029}\\ 
\footnotesize{{\bf USPS Digits} (multiclass)} &  0.3028  &  0.3377  &  0.4072  &  0.4821  &  0.4304  &  {\bf 0.5283}
\\
\bottomrule
\end{tabular}
\caption{Adjusted Rand (AR) index for binary and multiclass clustering of {\bf Vehicle}, {\bf MIT Scene}, and {\bf USPS Digits} datasets using $80$, $160$, and $200$ constraints, respectively.}\label{tab:smallest_const}
\end{table}

\subsection{Relative Constraints vs.\ Pairwise Constraints}

We first demonstrate the performance of the different methods using relative and pairwise constraints. 
For each dataset, we consider two different experiments: 
($i$)
\emph{binary} in which each dataset is clustered into {\em two groups}, 
based on some predefined criterion, and 
($ii$)
\emph{multi-class} where for each dataset the clustering is performed 
with number of clusters being equal to {\em number of classes}. 
In the binary experiment, we aim to find a crude partitioning of the data,  
while in the multi-class experiment we seek a clustering at a finer granularity. 

The 2-class partitionings of our datasets required for the binary experiment are defined as follows: 
For the {\bf vehicle} dataset, 
we consider class $4$ as one group and the rest of the classes as the second group (an arbitrary choice).
For the {\bf MIT Scene} dataset, we perform a partitioning of the data into natural vs.\ man-made scenes. 
Finally, for the {\bf USPS Digits}, we divide the data instances into even vs.\ odd digits. 

To generate the pairwise constraints for each dataset, 
we vary the number of labeled instances from each class
(from $5$ to $19$ with step-size of $2$) 
and form all possible ML constraints. 
We then consider the same number of CL constraints.
Note that for the binary case, we only have two classes for each dataset. 
To compare with the methods that use relative comparisons, 
we consider an equal number of relative comparisons 
and generate them by sampling two random points from the same class and 
one point (outlier) from one of the other classes.
Note that for the relative comparisons, there is no need to restrict the points to the labeled samples, as the comparison is made in a relative manner.

One should note the fundamental difference between the pairwise and relative constraints when comparing the different methods using different types of constraints. At first glance, the number of  independent pairwise comparisons for detecting the outlier among three given items might seem larger than the one in a single pairwise comparison task (ML or CL). However, one should also consider the contrast imposed by the existence of the two similar items which facilitates the decision making task for the user. In other words, we argue that the user does not perform three independent pairwise comparisons for detecting the outlier. Additionally, we allow the user to skip the difficult cases but use the unspecified answers to improve the final results. In short, we acknowledge that it might not be totally plausible to compare different methods that use different types of distance constraints. However, the comparison provides insightful intuition to assess the performance of these methods in different settings.

Finally, in these experiments, we consider a subsample of both 
{\bf MIT Scene} and {\bf USPS Digits} datasets 
by randomly selecting $100$ data points from each class, 
yielding $800$ and $1000$ data points, respectively.

\setcounter{subfigure}{-8}
\begin{figure*}[t]
	\begin{center}
	\subfigure{\includegraphics[width=0.24\textwidth]{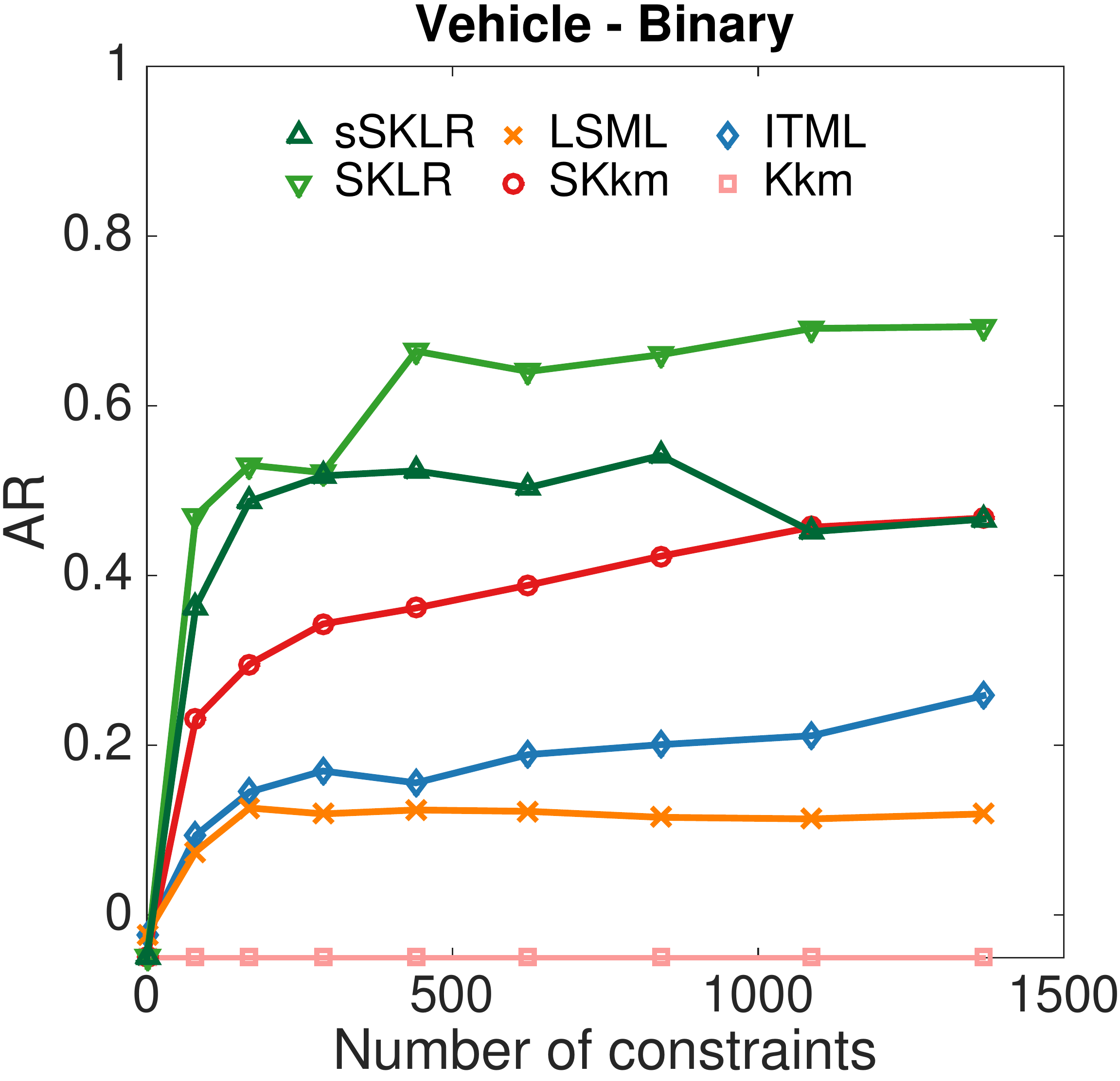}}\hfill
	\subfigure{\includegraphics[width=0.24\textwidth]{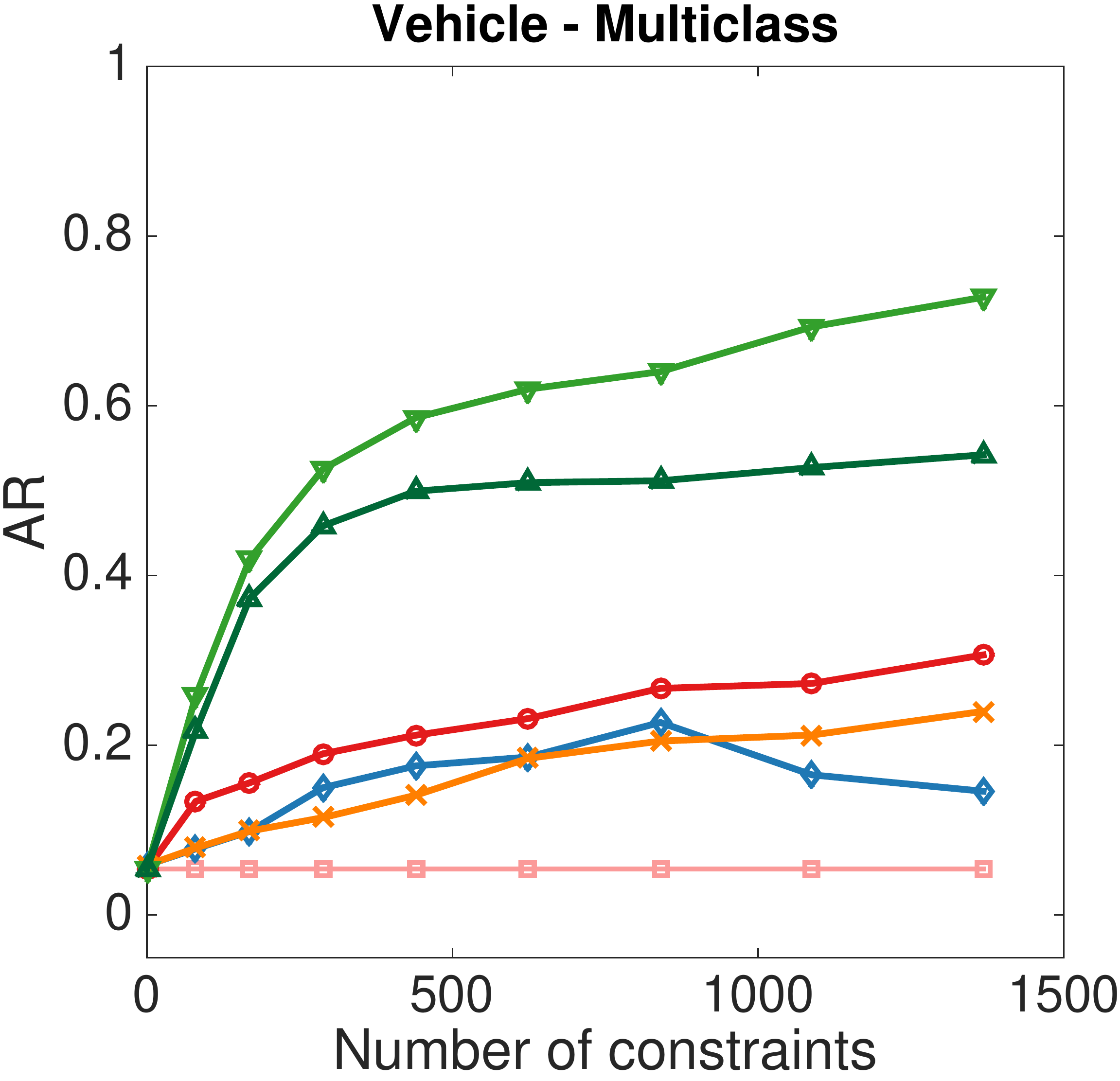}}\hfill
	\subfigure{\includegraphics[width=0.24\textwidth]{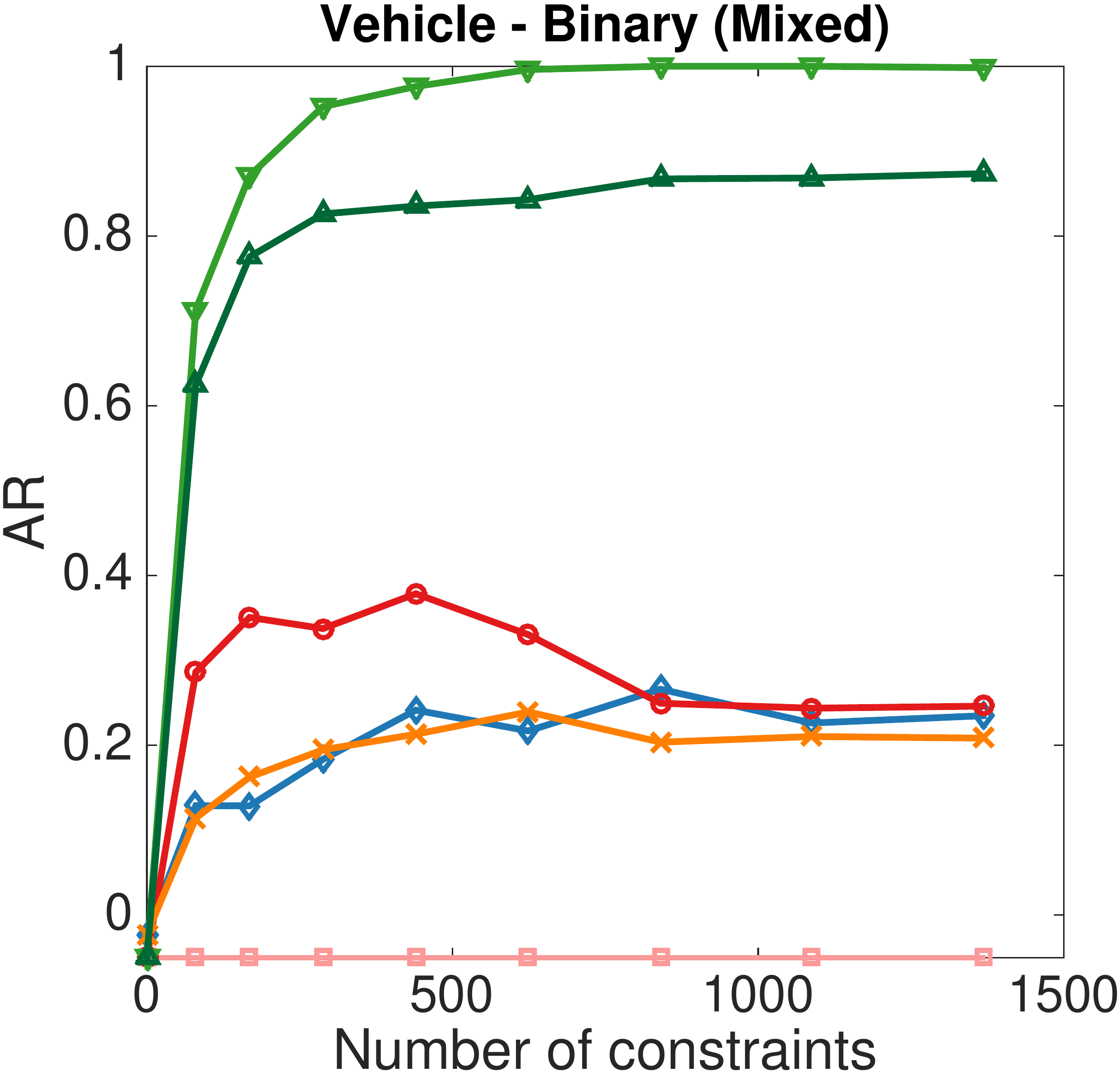}}\hfill
	\subfigure{\includegraphics[width=0.24\textwidth]{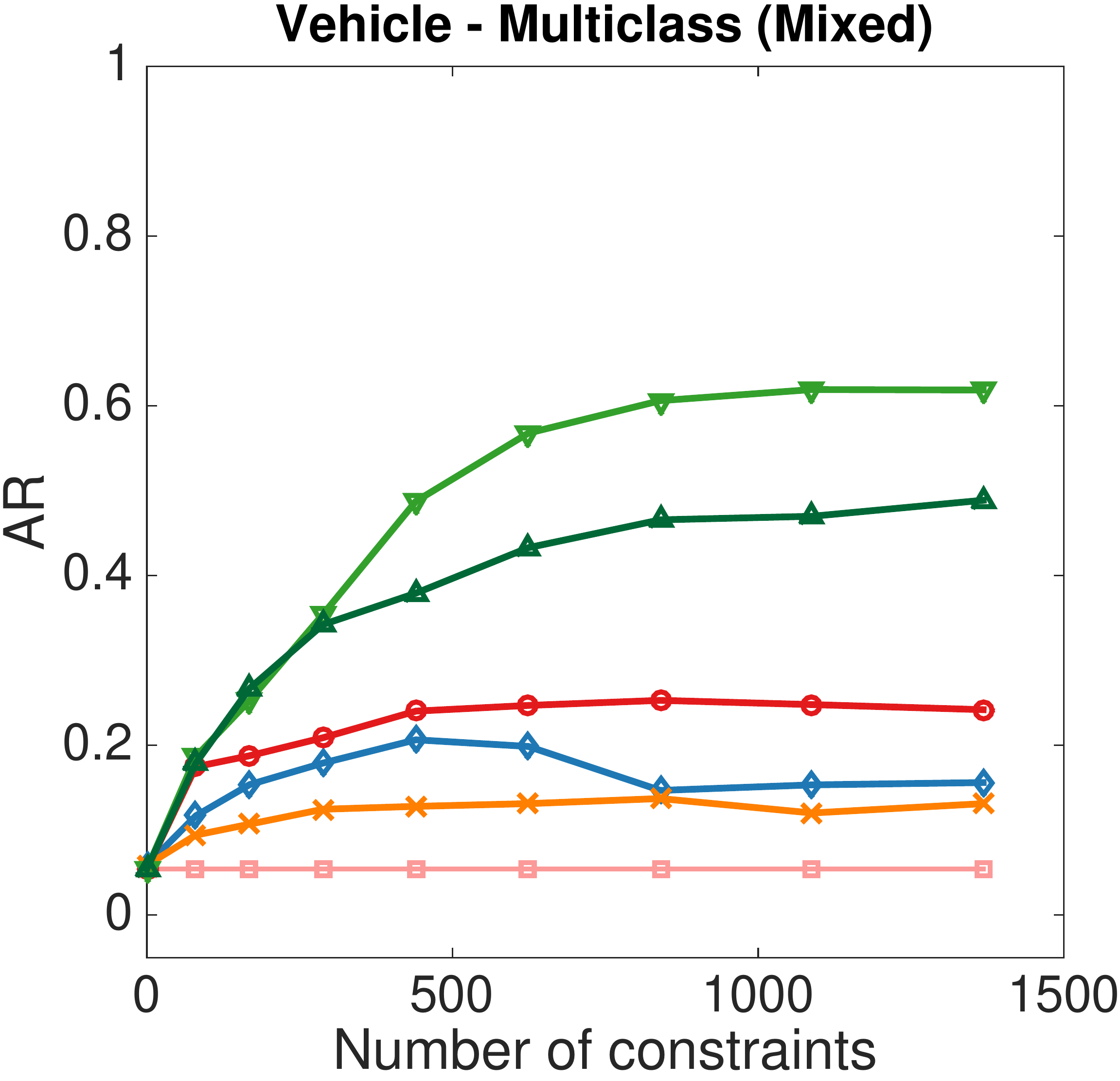}}\hfill\\
	\subfigure{\includegraphics[width=0.24\textwidth]{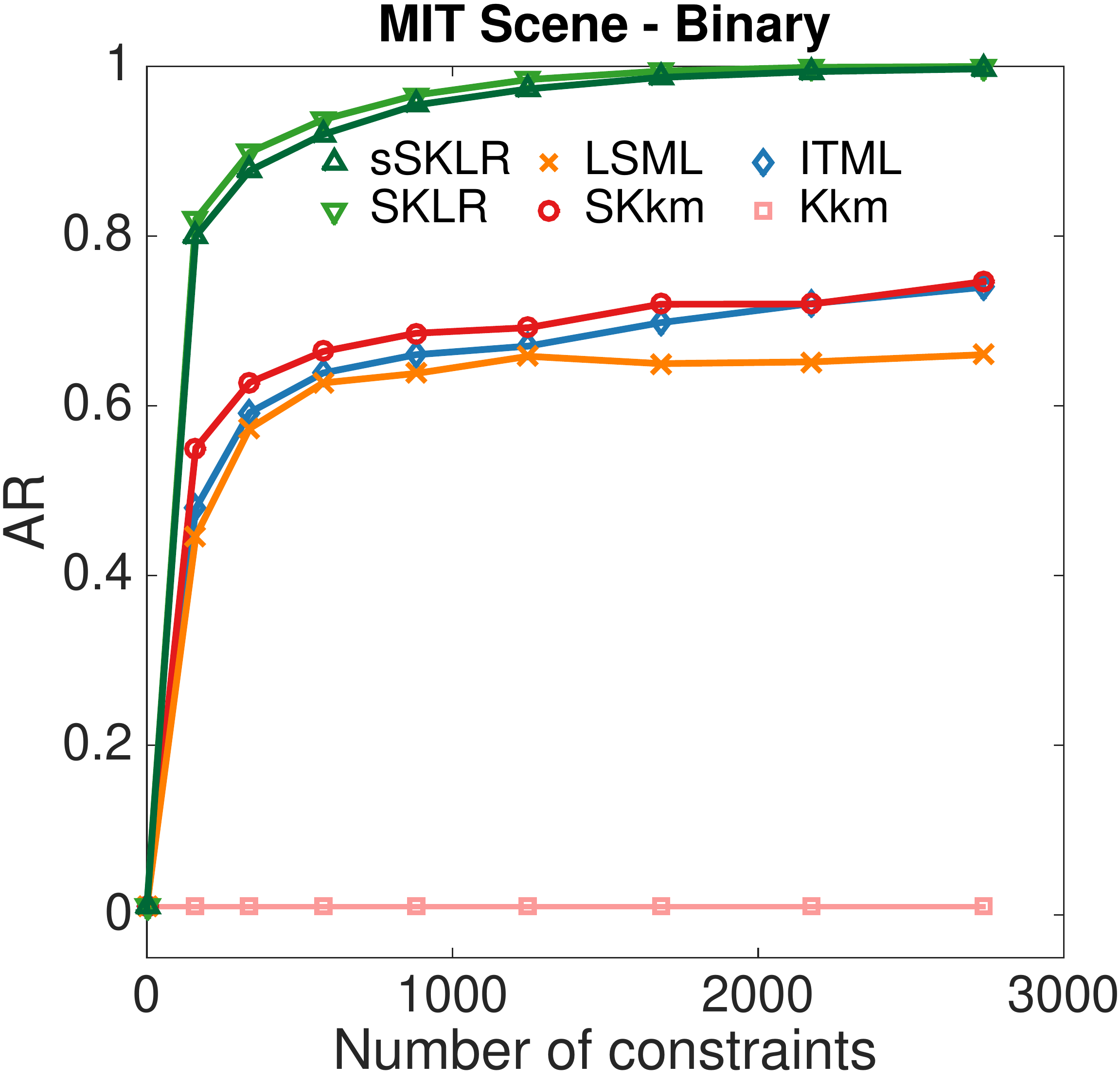}}\hfill
	\subfigure{\includegraphics[width=0.24\textwidth]{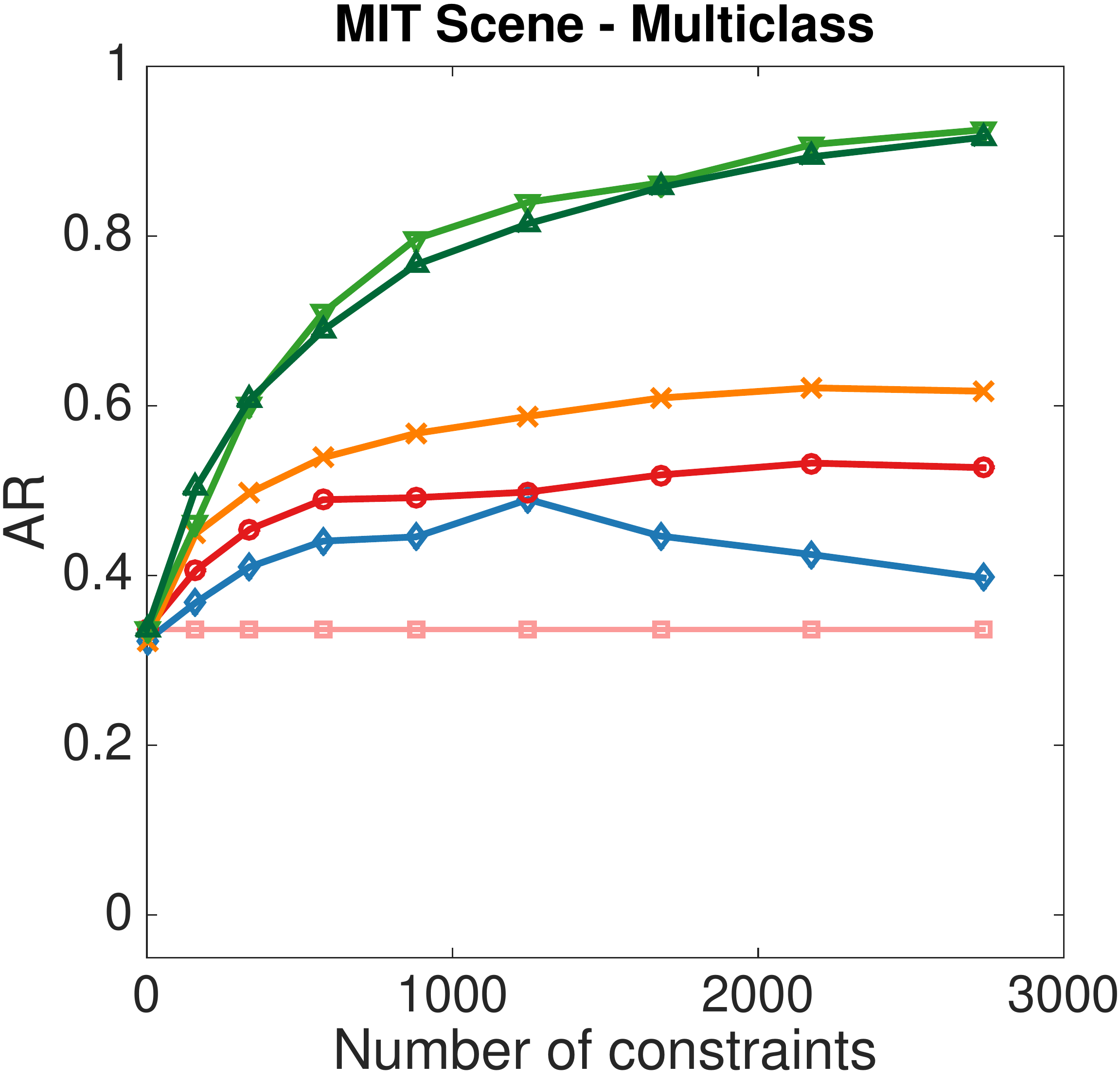}}\hfill
	\subfigure{\includegraphics[width=0.24\textwidth]{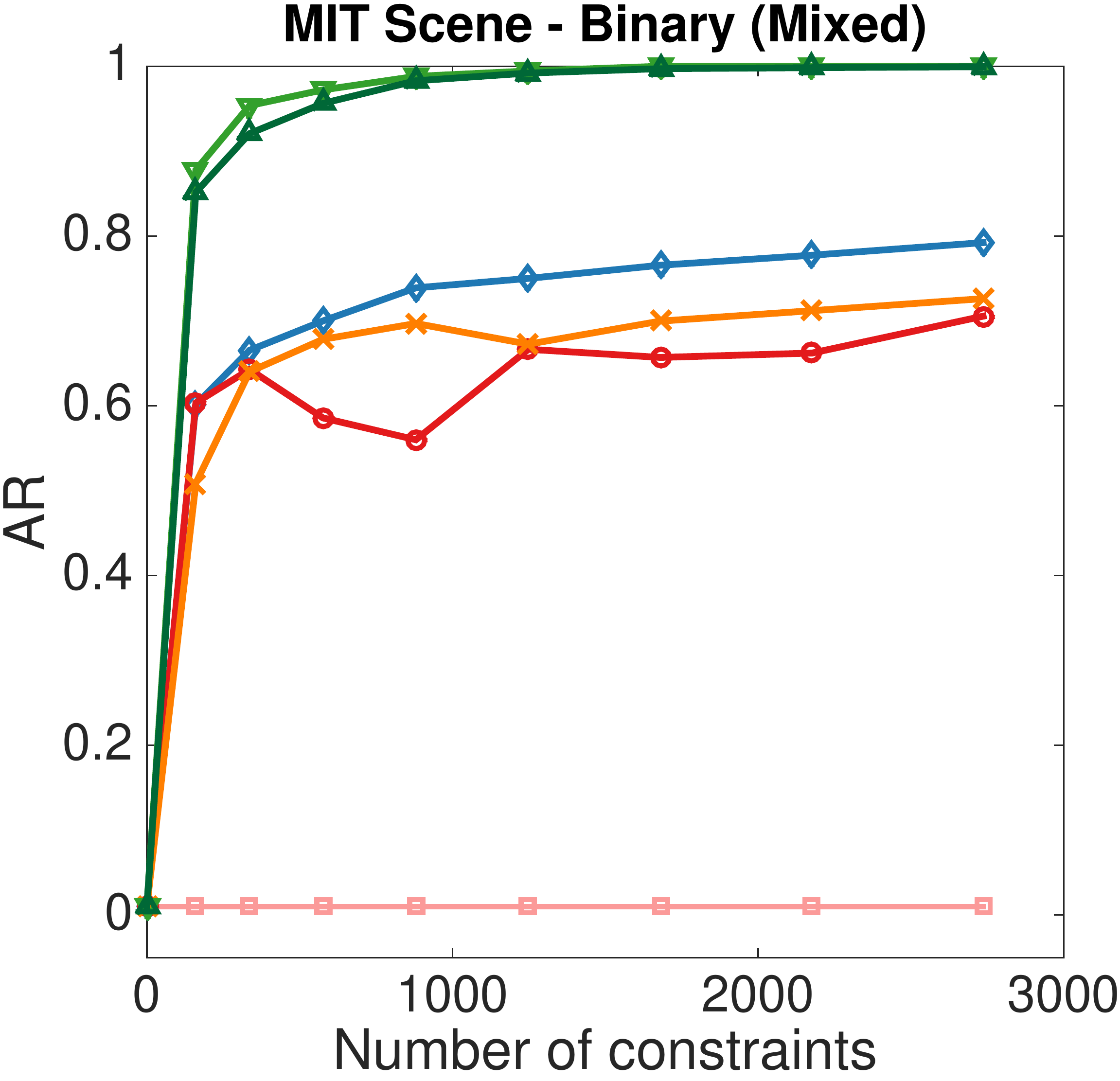}}\hfill
	\subfigure{\includegraphics[width=0.24\textwidth]{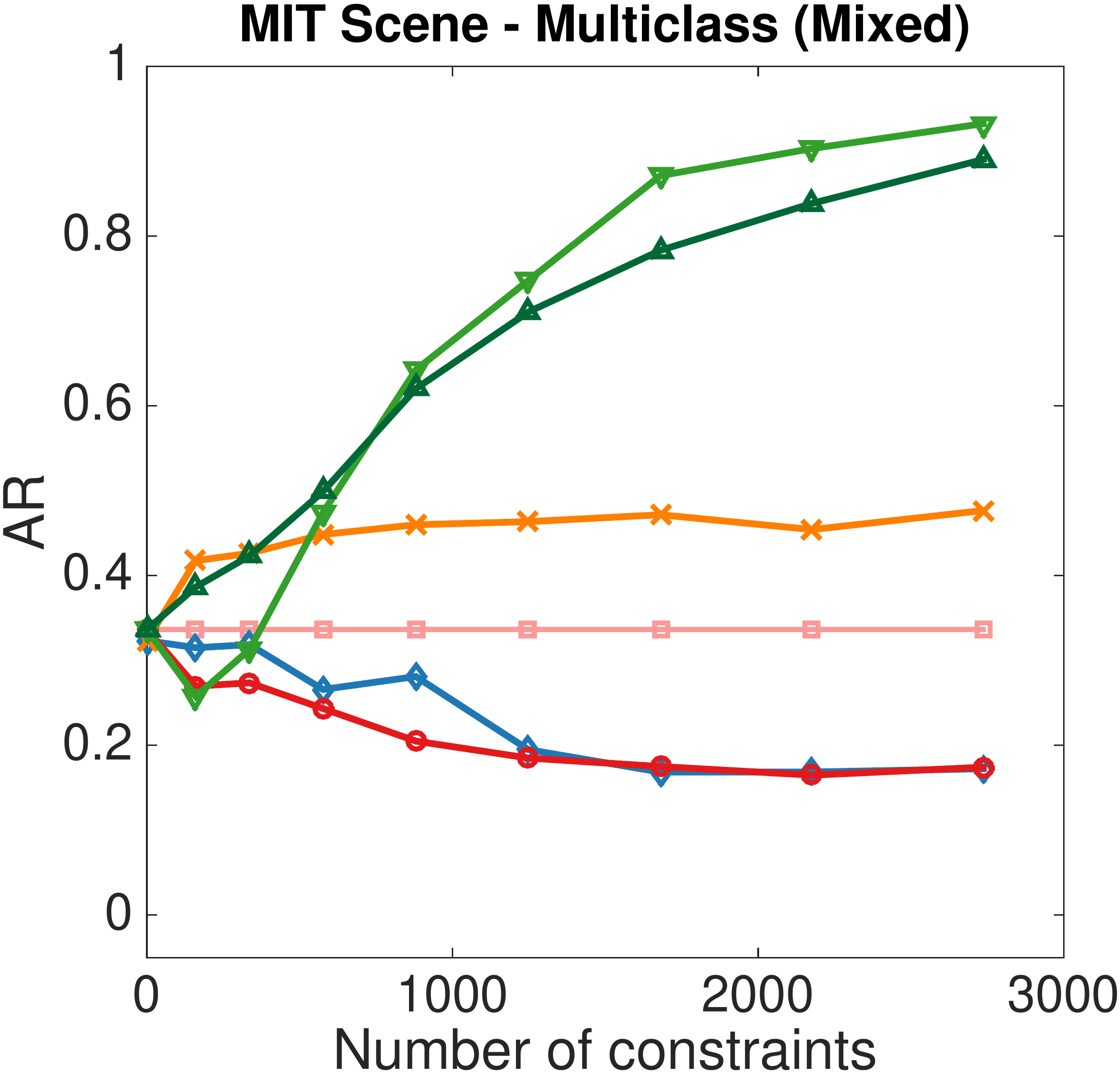}}\hfill\\
	\subfigure[]{\includegraphics[width=0.24\textwidth]{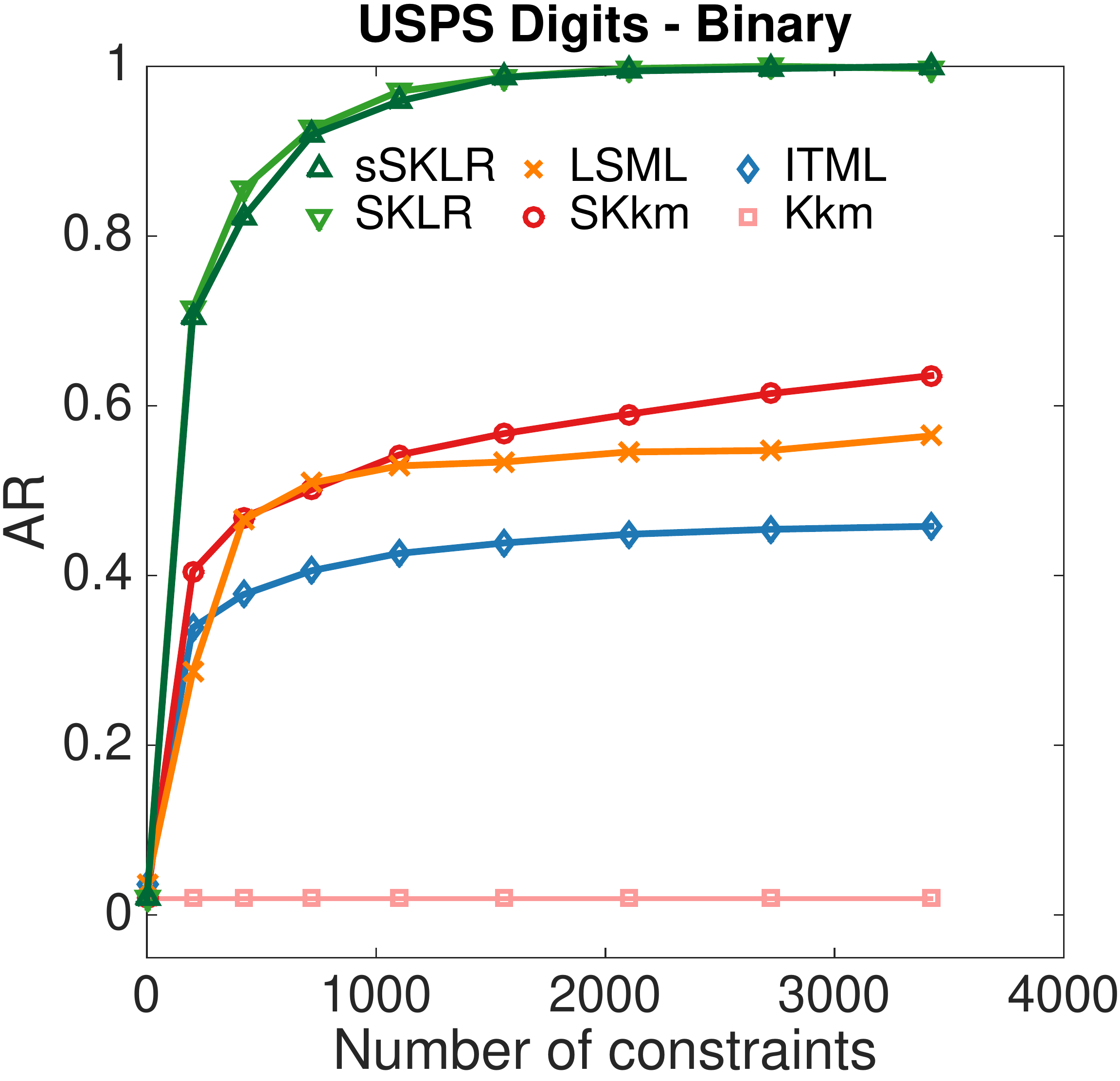}\label{fig:binary}}\hfill
	\subfigure[]{\includegraphics[width=0.24\textwidth]{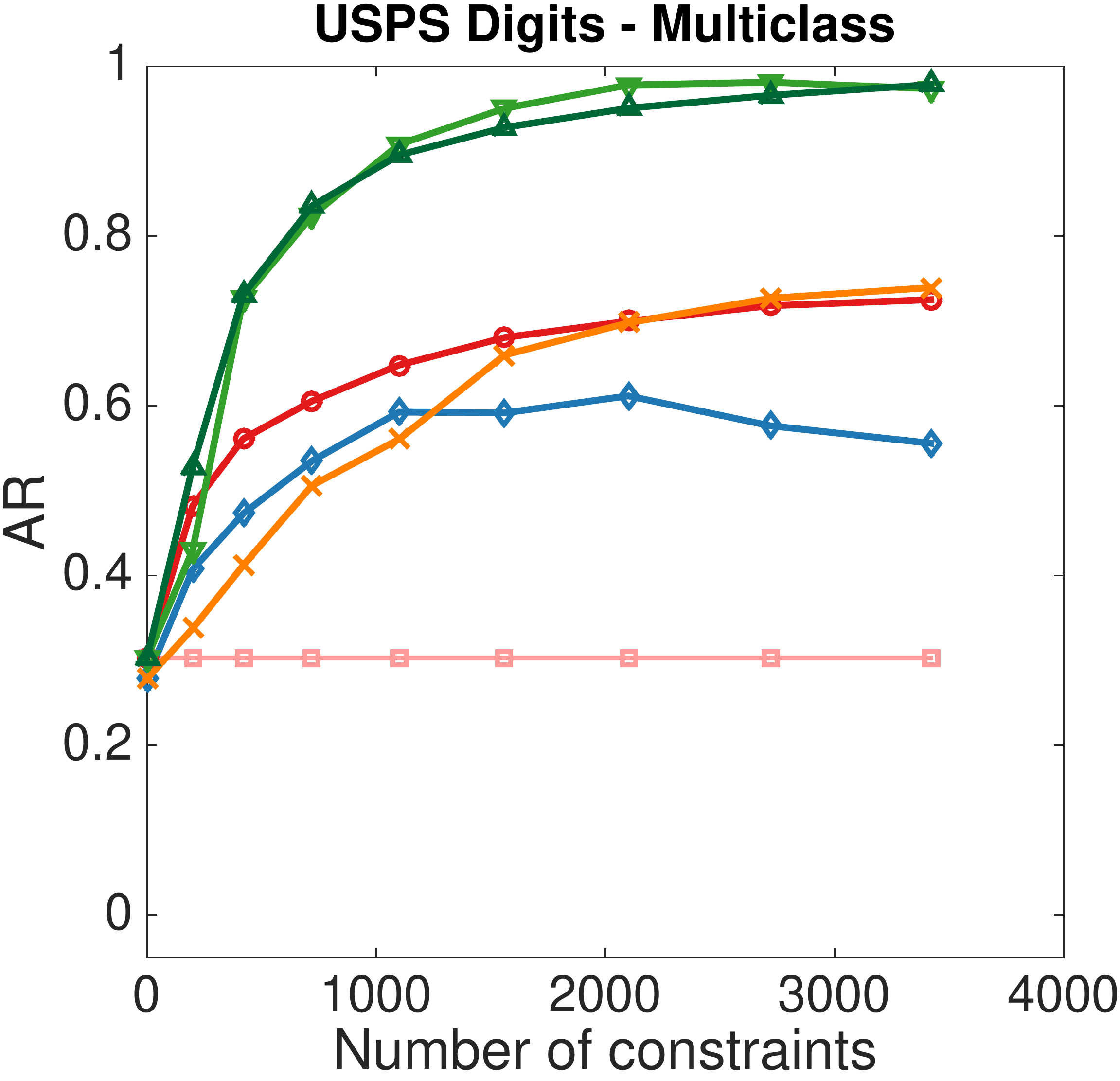}\label{fig:multiclass}}\hfill
	\subfigure[]{\includegraphics[width=0.24\textwidth]{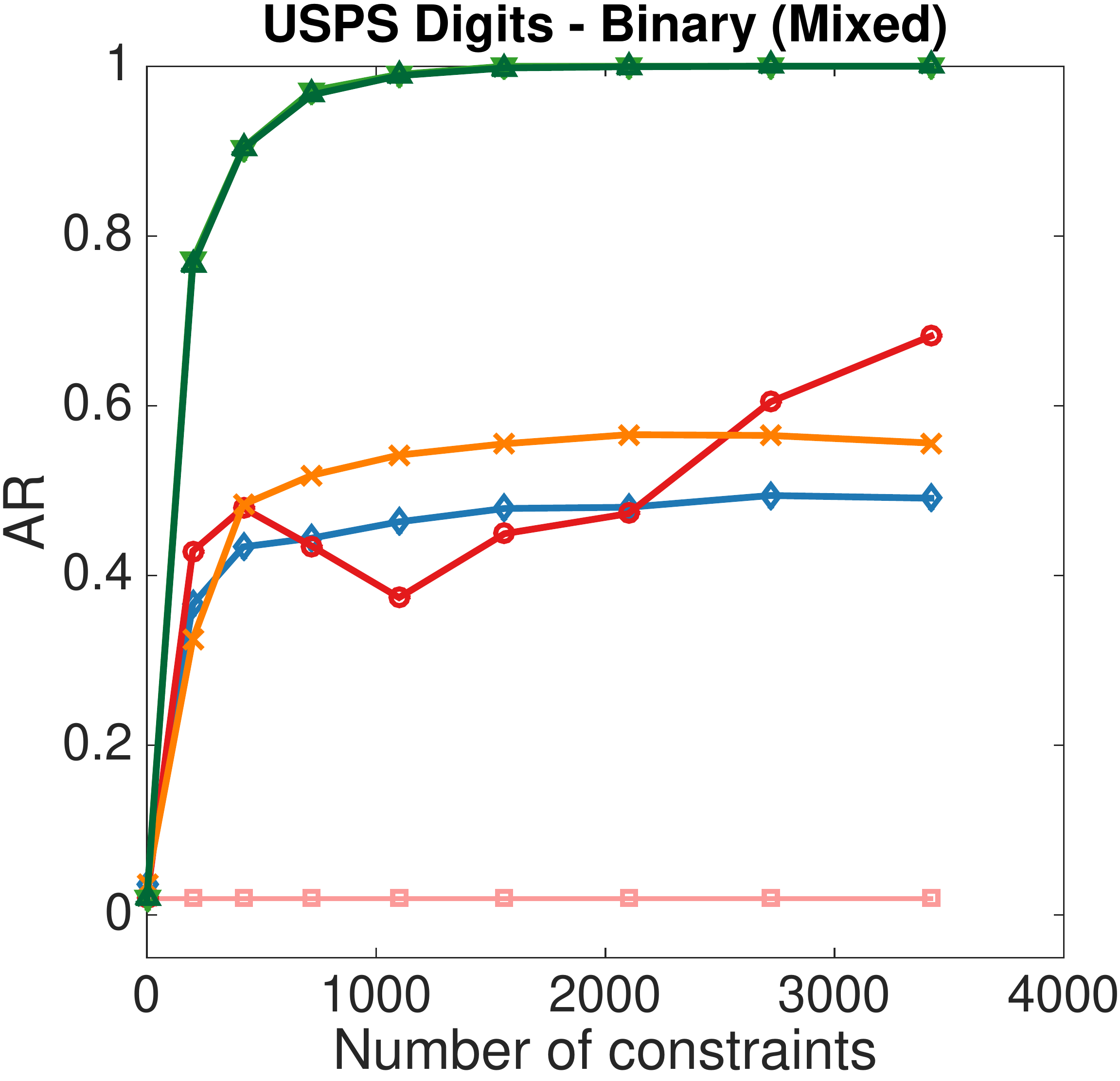}\label{fig:binary_mixed}}\hfill
	\subfigure[]{\includegraphics[width=0.24\textwidth]{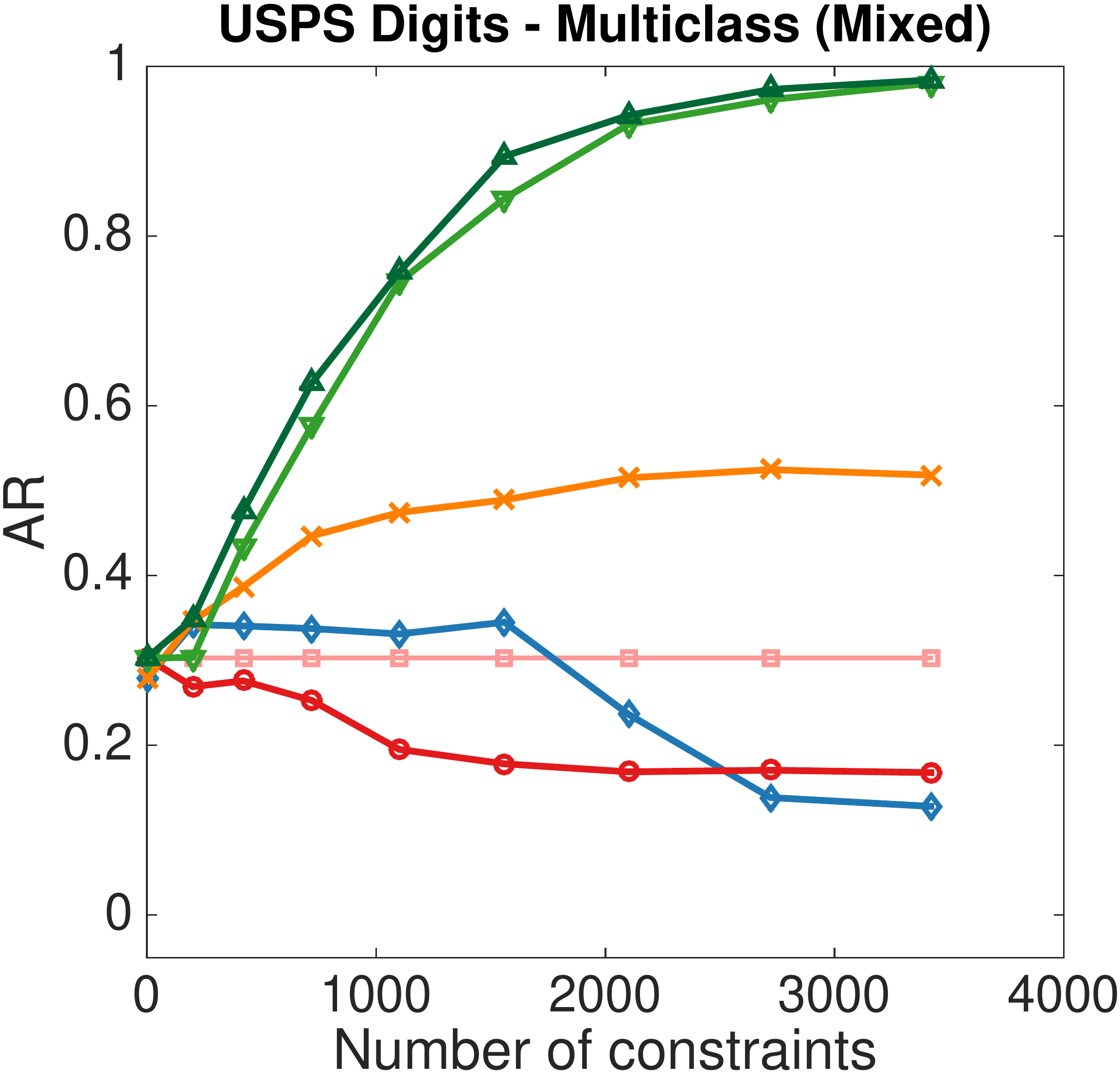}\label{fig:multiclass_mixed}}\hfill
\caption{\label{fig:clustering_results}
Clustering accuracy measured with Adjusted Rand index (AR). 
Rows correspond to different datasets:
(1)~{\bf Vehicle};
(2)~{\bf MIT Scene};
(3)~{\bf USPS Digits}. 
Columns correspond to different experimental settings: 
(a)~binary with  separate constraints;
(b)~multi-class  with separate constraints;
(c)~binary  with mixed constraints;
(d)~multi-class with mixed constraints.}
\end{center}
\end{figure*} 


The results for the binary and multi-class experiments 
are shown in Figures~\ref{fig:binary} and~\ref{fig:multiclass}, respectively. 
We see that all methods perform equally with no constraints.
As constraints or relative comparisons are introduced
the accuracy of all methods improves very rapidly.
The only surprising behavior is the one of {\sf ITML} in the multi-class setting, 
whose accuracy drops as the number of constraints increases. Table~\ref{tab:smallest_const} shows the  performance of different methods for binary and multiclass clustering of the datasets after introducing constraints computed from five labeled examples per class.
(This is the least amount of supervision we envision is feasible in practical settings.)
As can be seen, the increase in performance is much larger for both {\sf SKLR} and {\sf sSKLR} methods in all cases.
From the figures we see that {\sf SKLR} outperforms all competing methods by a large margin, 
for all three datasets and in both settings. Additionally, the overall performance of {\sf sSKLR} is only slightly lower than {\sf SKLR}, however, it still outperforms all the other methods. 

\begin{figure*}[t]
	\begin{center}
	\subfigure[]{\includegraphics[width=0.32\textwidth]{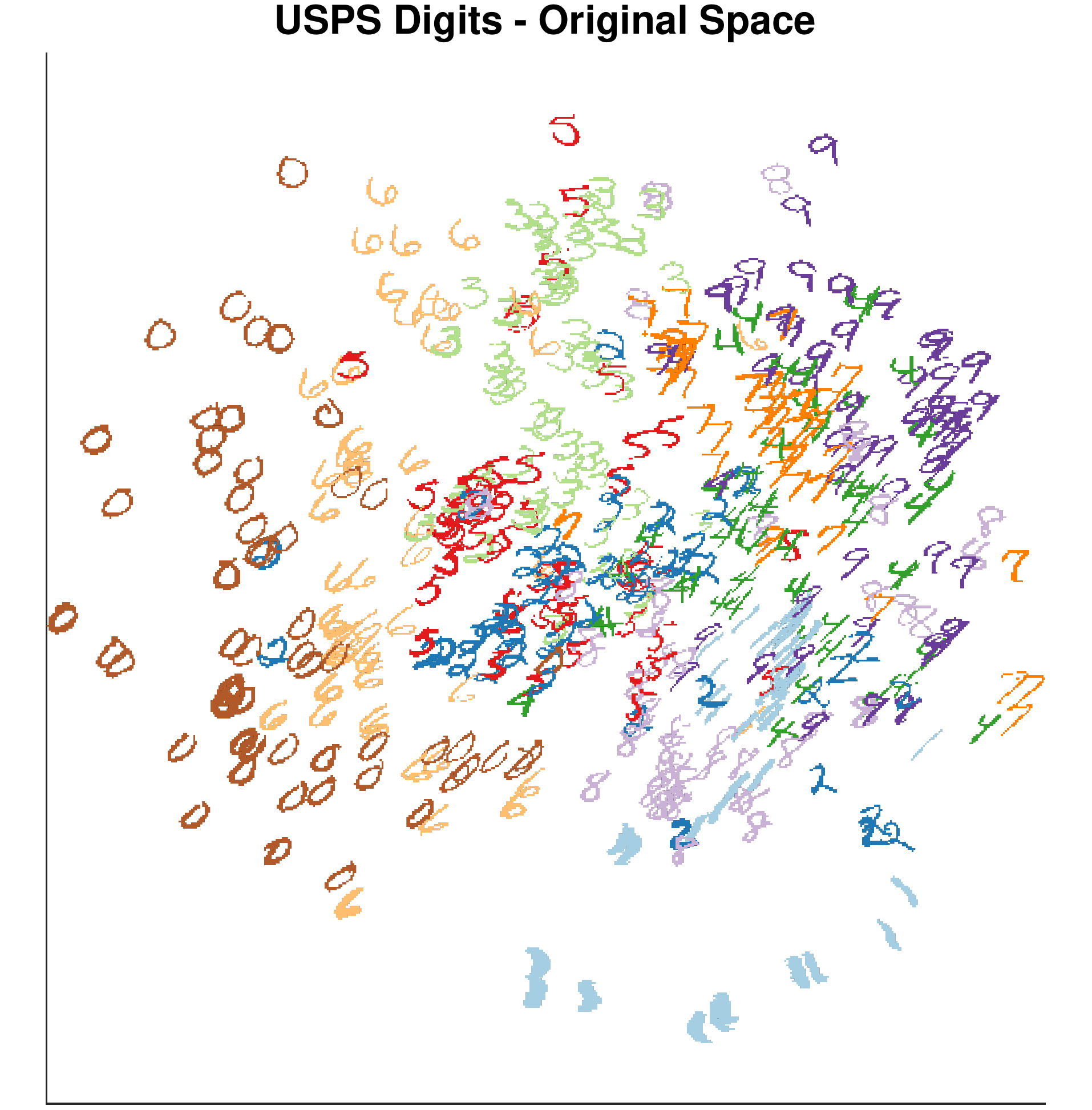}\label{fig:usps_orig}}
	\subfigure[]{\includegraphics[width=0.32\textwidth]{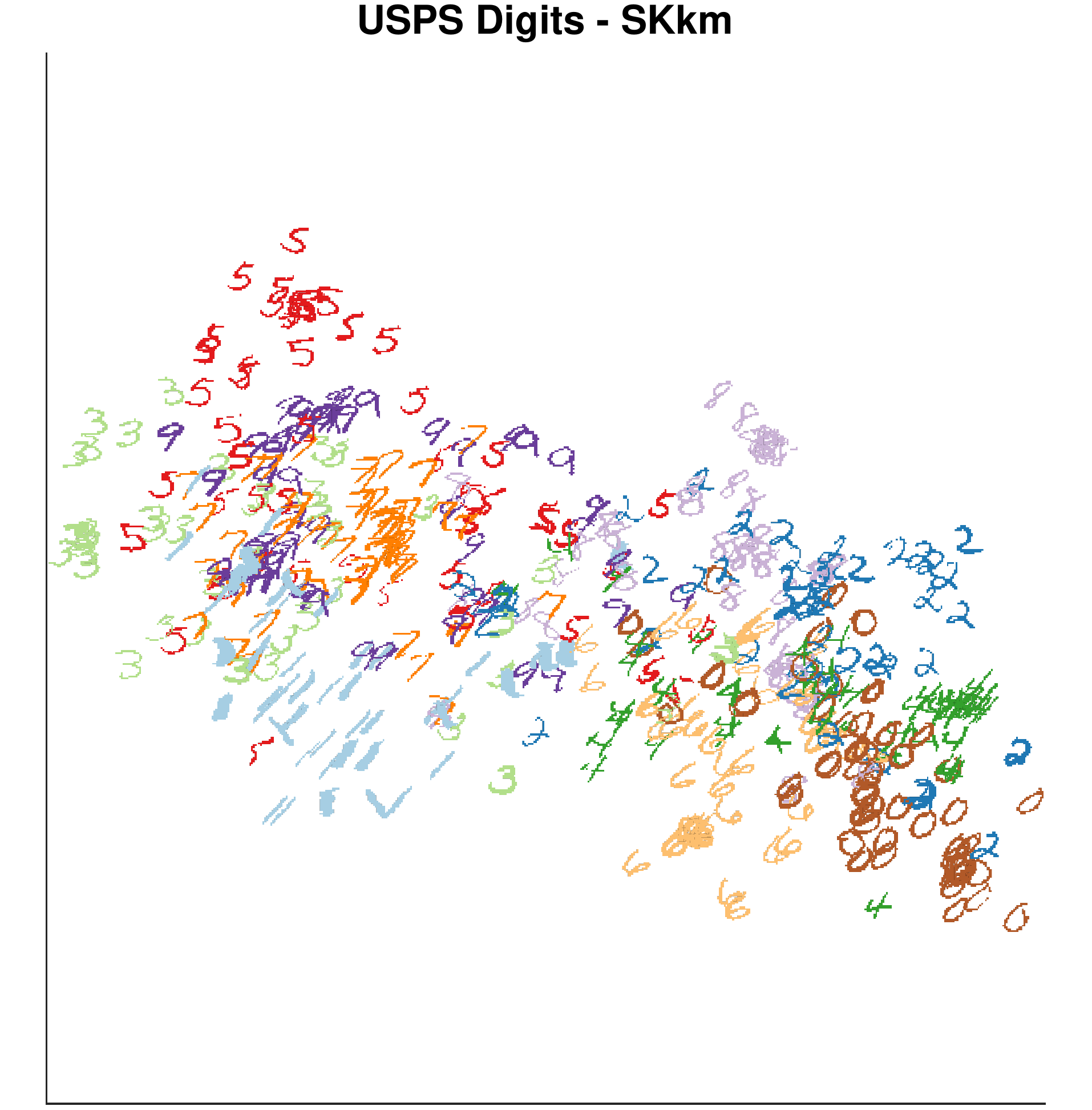}\label{fig:usps_skms}}
	\subfigure[]{\includegraphics[width=0.32\textwidth]{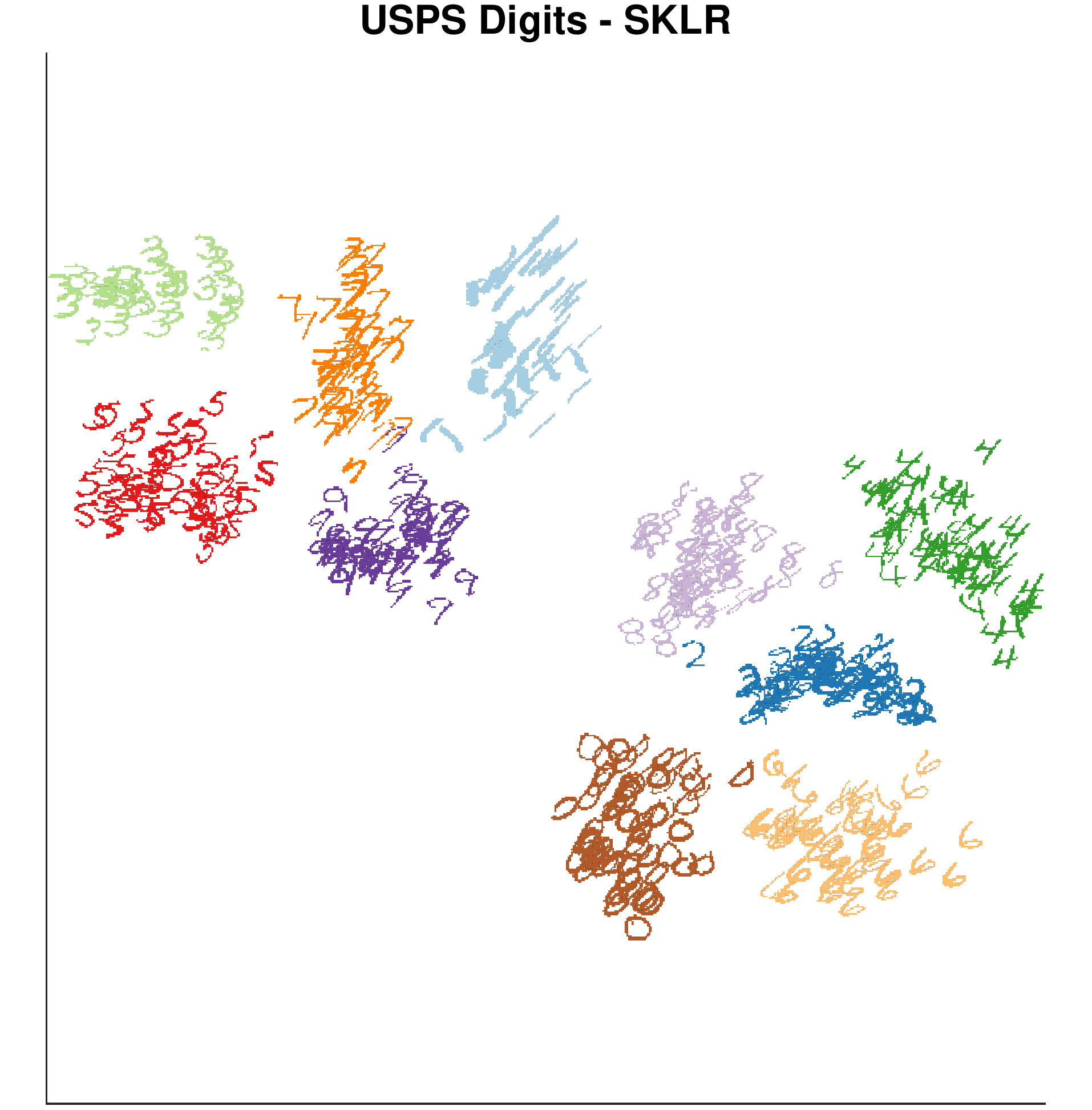}\label{fig:usps_proposed}}
	\caption{\label{fig:usps_vis}Visualization of the {\bf USPS Digits} using SNE: 
	(a) original space; (b) space obtained by {\sf SK$k$m}; 
	(c) space obtained by our method, {\sf SKLR}. Note the clusters in two different resolutions: even vs. odd (binary), and digits (multiclass).}
\end{center}
\end{figure*} 

\setcounter{subfigure}{-4}
\begin{figure*}[t]
	\begin{center}
	\subfigure{\includegraphics[width=0.24\textwidth]{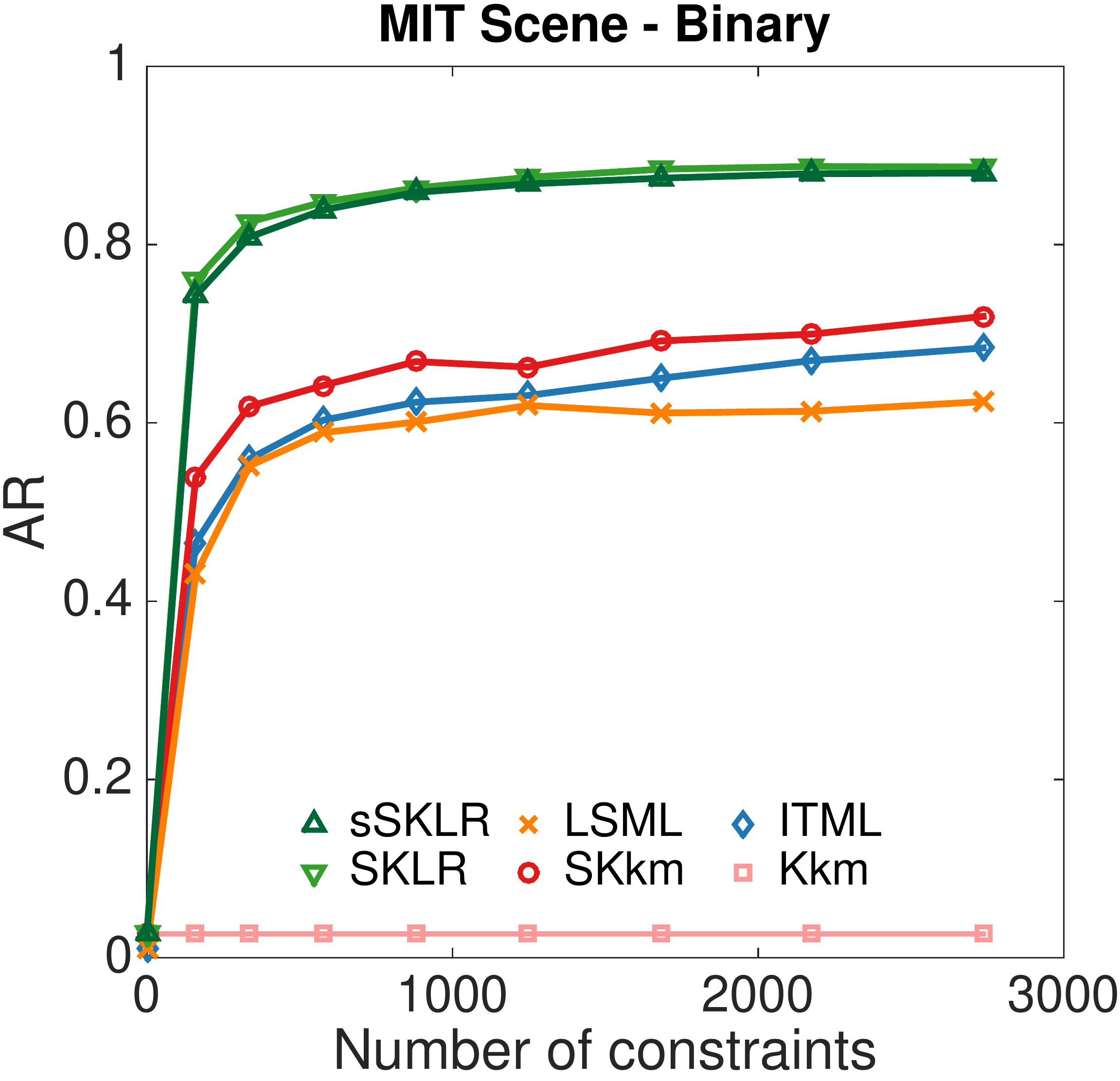}}\hfill
	\subfigure{\includegraphics[width=0.24\textwidth]{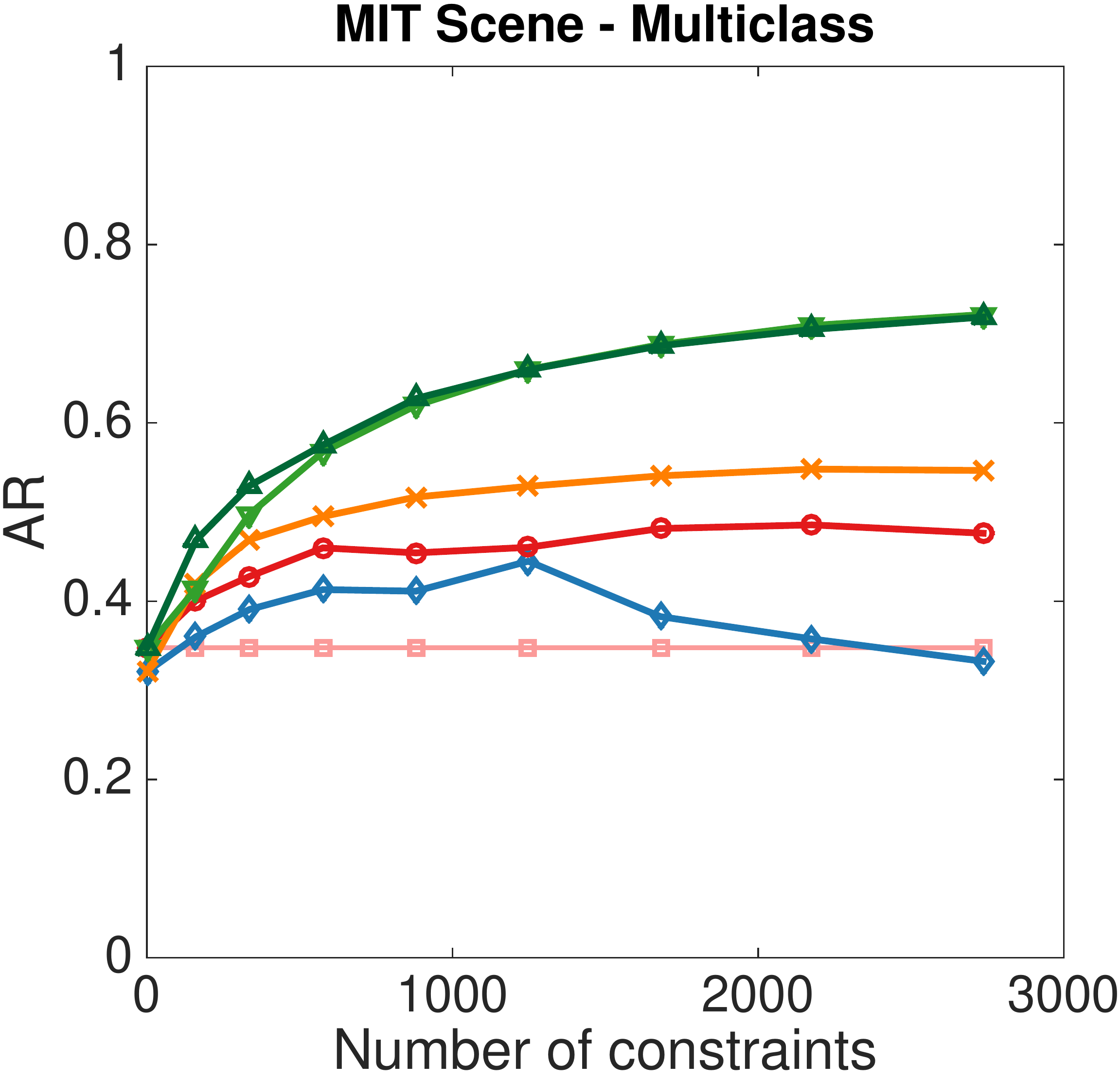}}\hfill 
	\subfigure{\includegraphics[width=0.24\textwidth]{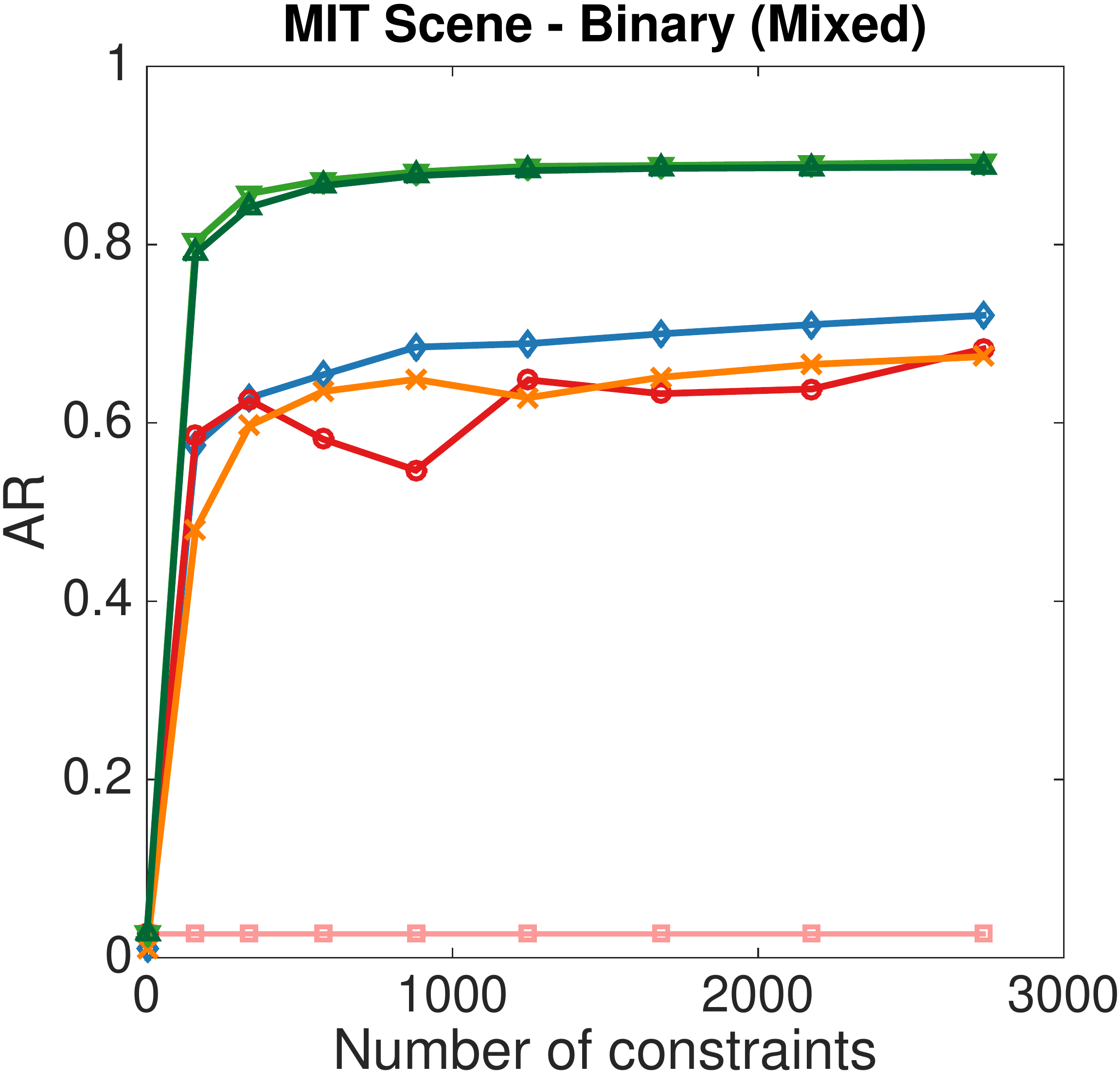}}\hfill
	\subfigure{\includegraphics[width=0.24\textwidth]{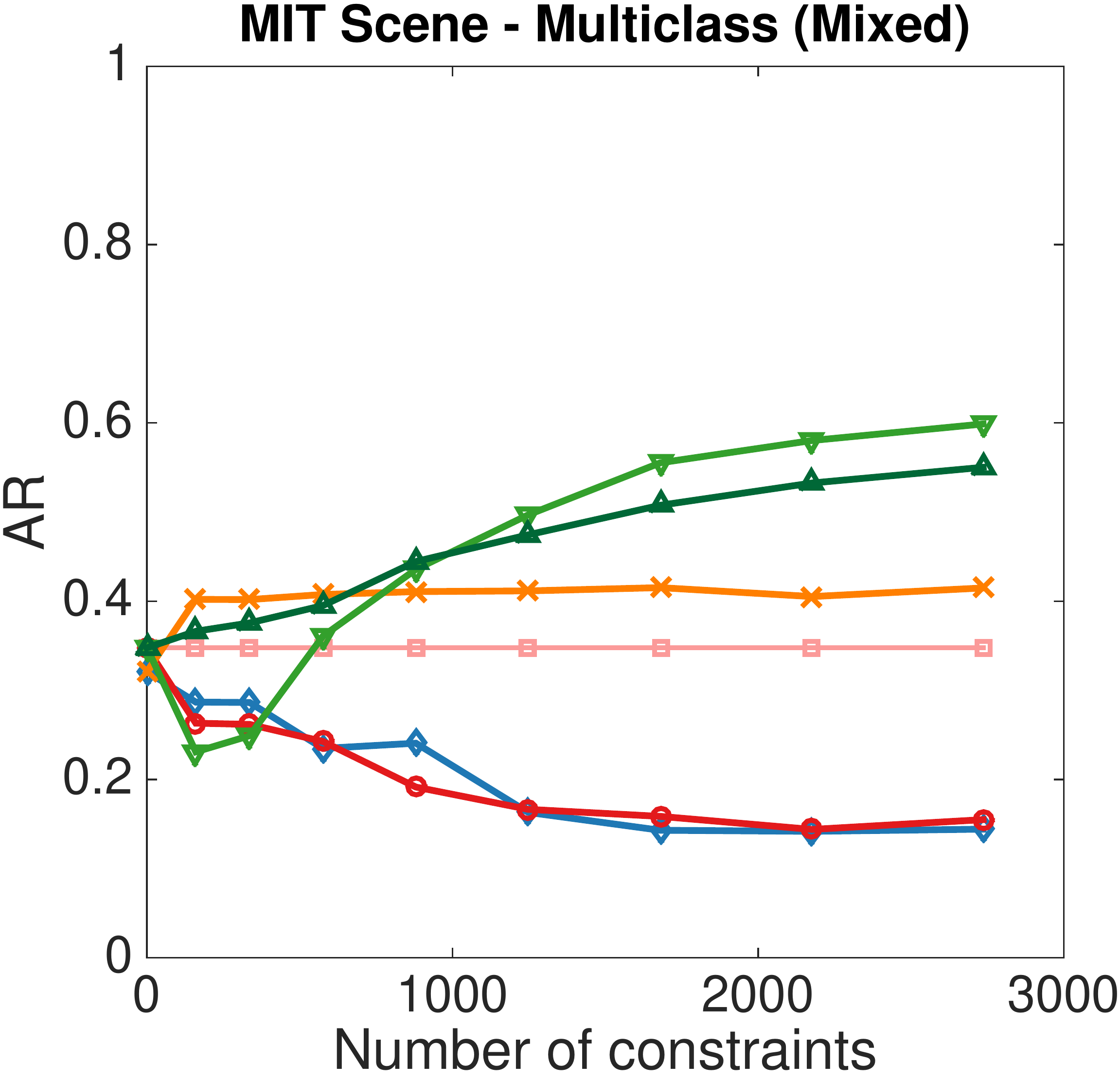}}\hfill\\
	\subfigure[]{\includegraphics[width=0.24\textwidth]{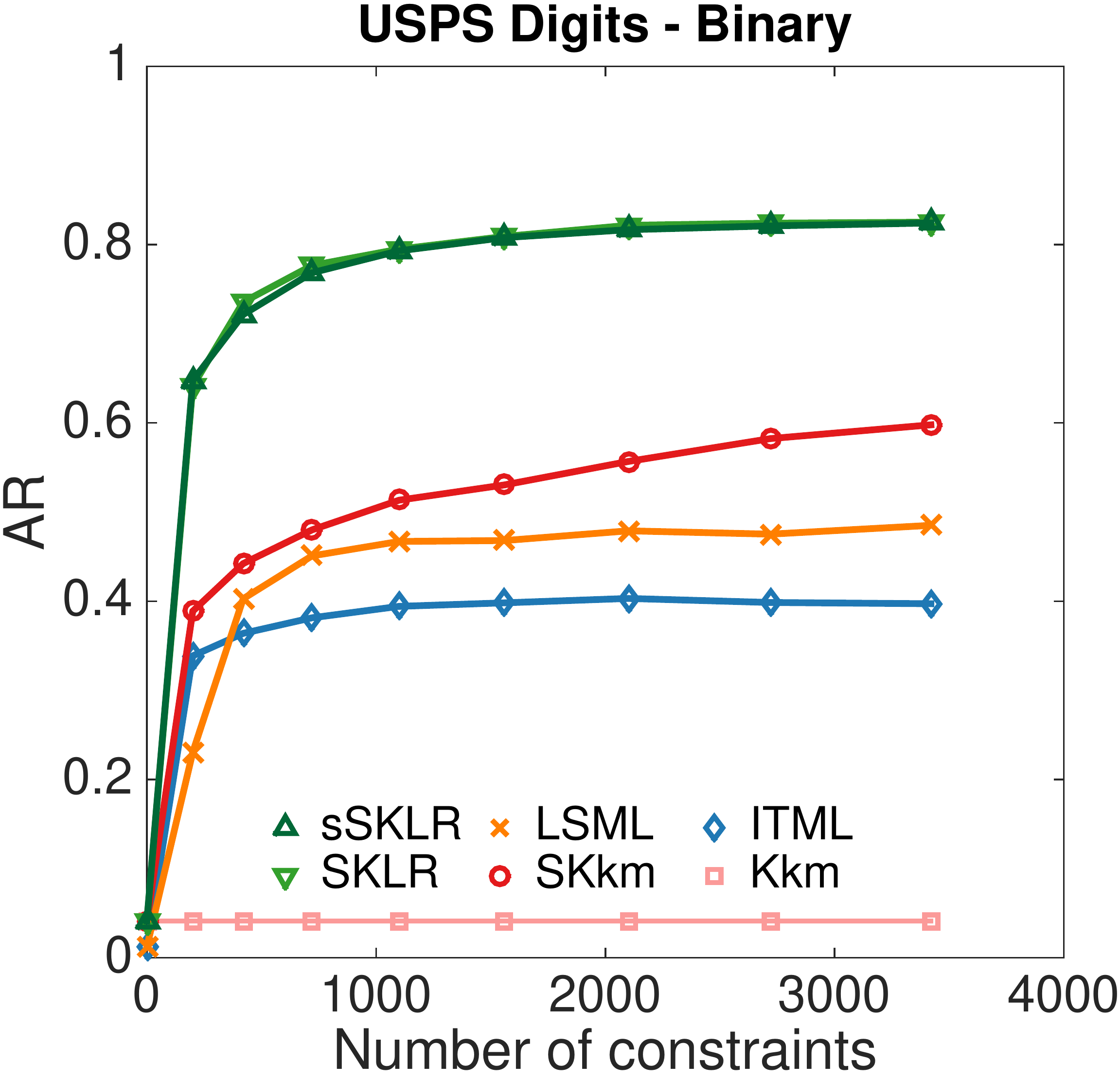}\label{fig:usps_multi_mixed}}\hfill
	\subfigure[]{\includegraphics[width=0.24\textwidth]{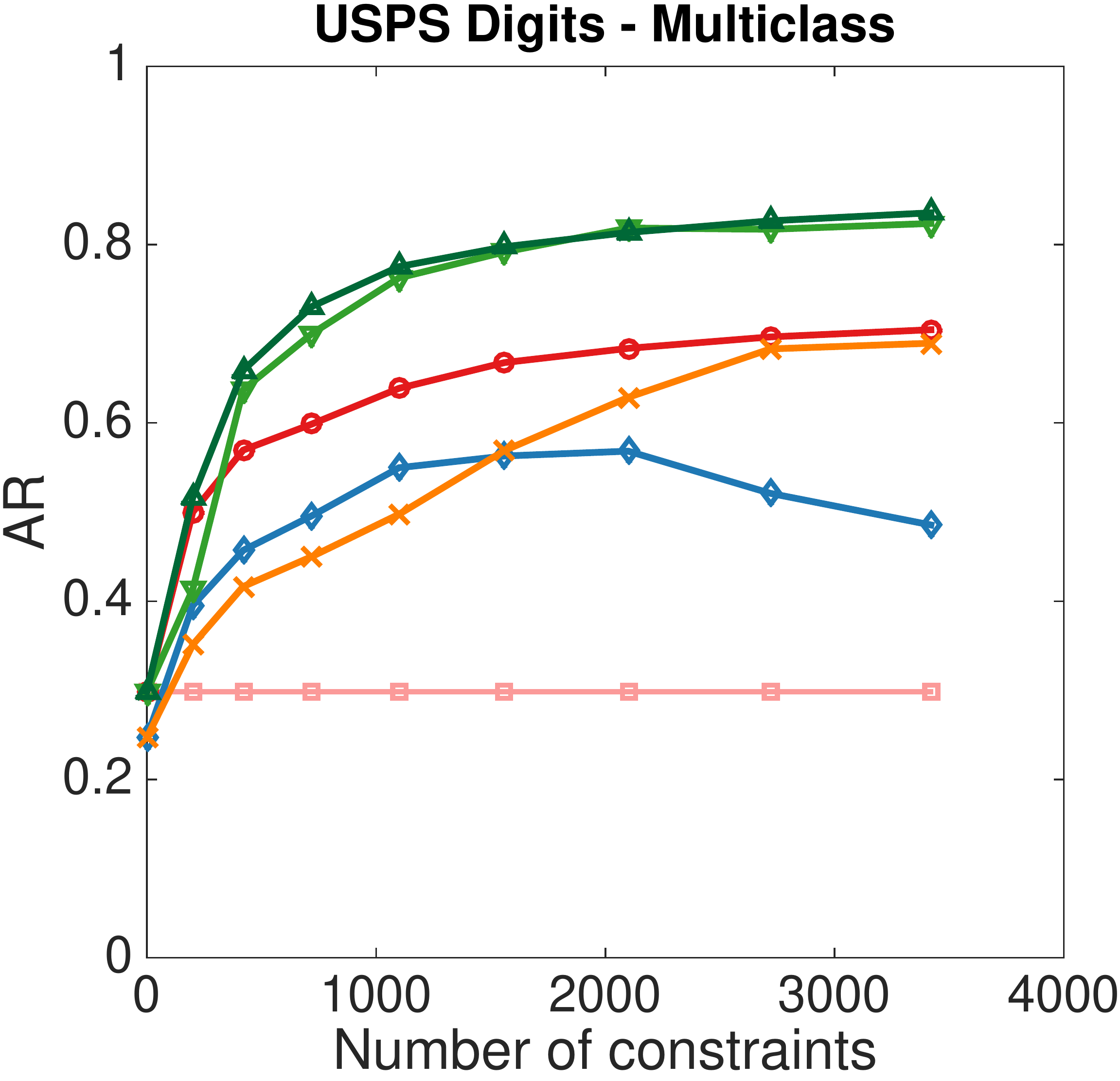}\label{fig:usps_bin_gen}}\hfill
	\subfigure[]{\includegraphics[width=0.24\textwidth]{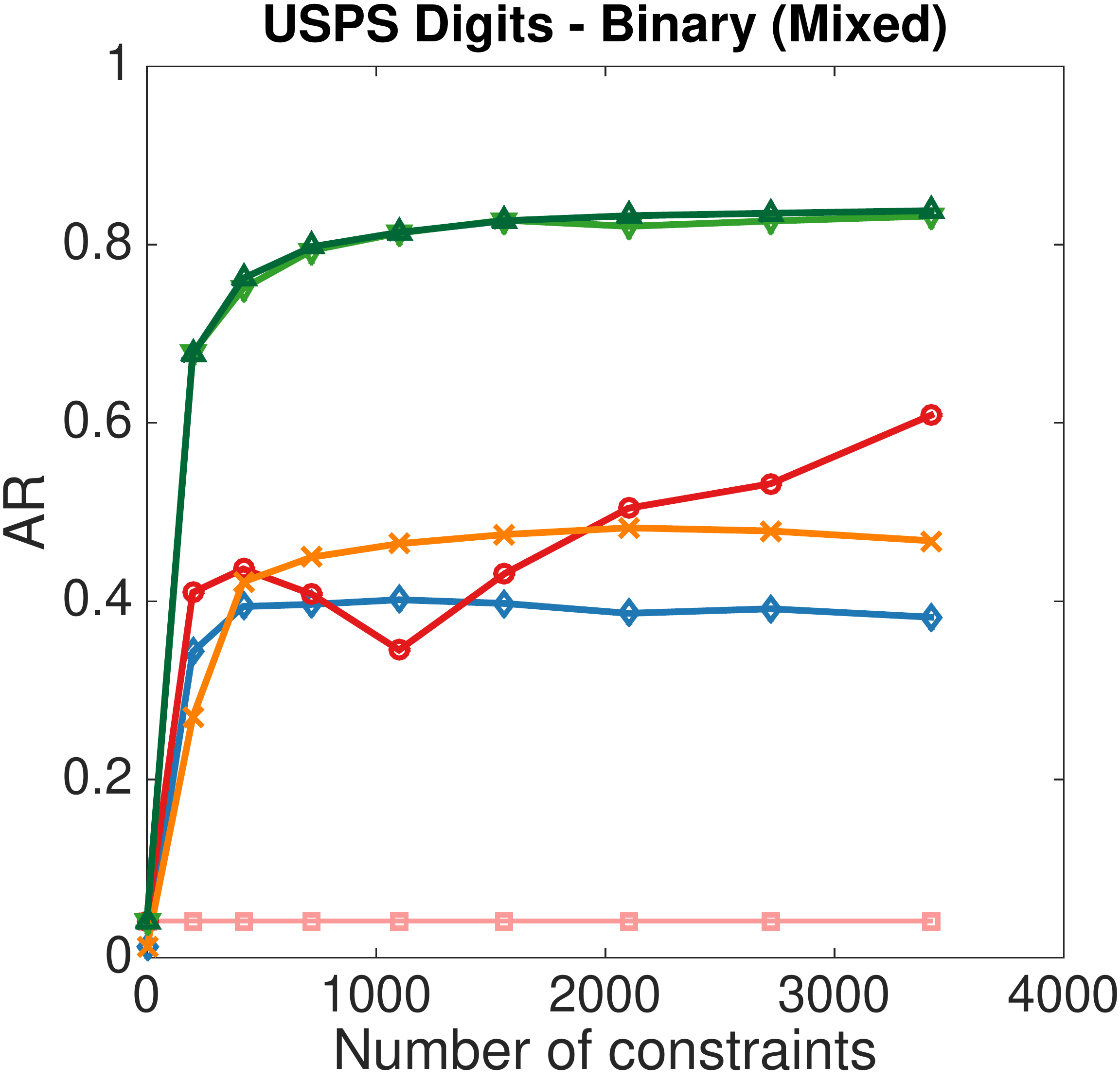}\label{fig:usps_multi_gen_mixed}}\hfill
	\subfigure[]{\includegraphics[width=0.24\textwidth]{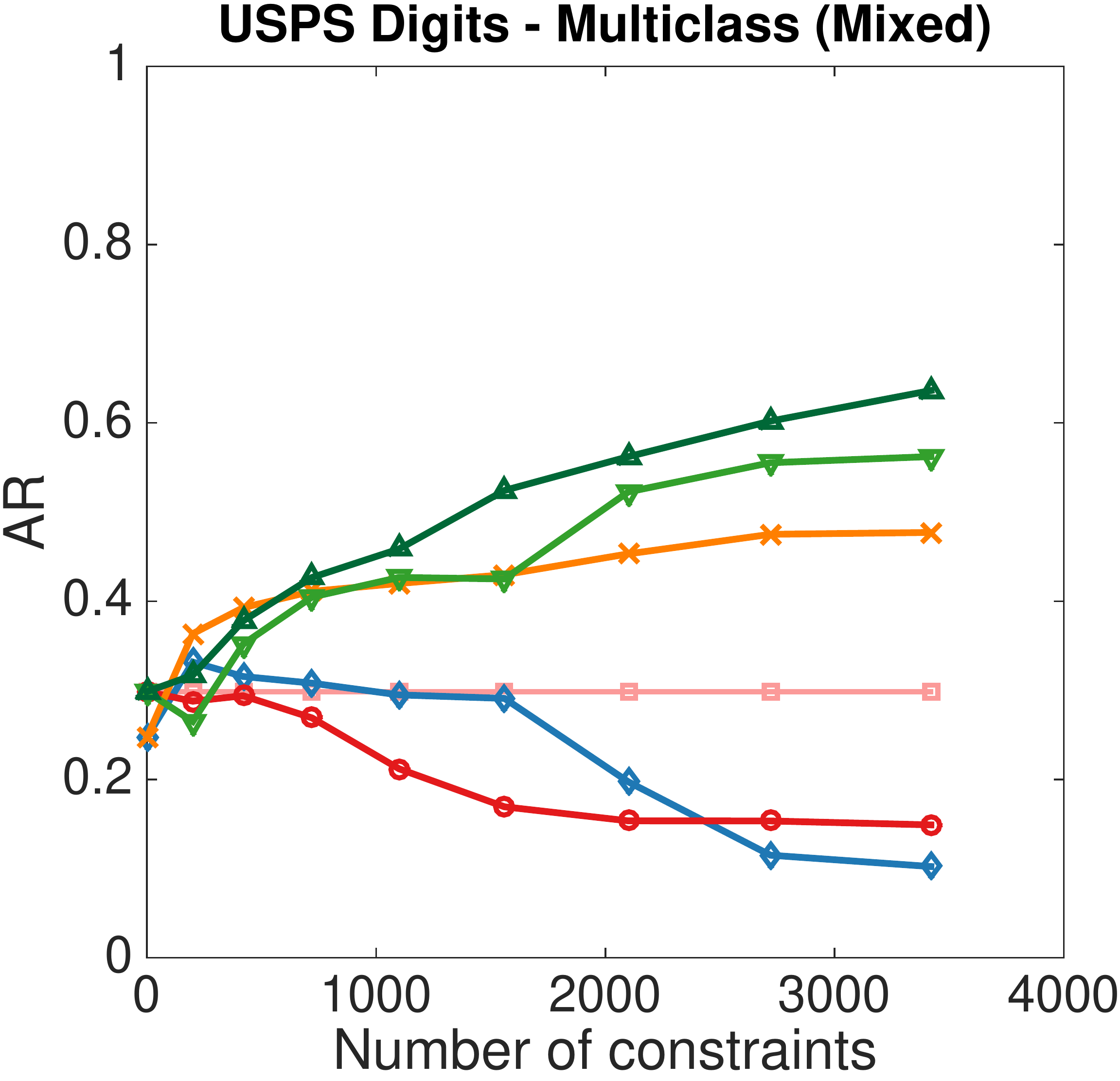}\label{fig:usps_bin_gen_mixed}}\hfill\\
	\caption{\label{fig:generalization_results}
Clustering accuracy on out-of-sample data (generalization performance). 
Rows correspond to different datasets:
(1)~{\bf MIT Scene};
(2)~{\bf USPS Digits}. 
Columns correspond to different experimental settings: 
(a)~binary with  separate constraints;
(b)~multi-class  with separate constraints;
(c)~binary  with mixed constraints;
(d)~multi-class with mixed constraints.	
}

\end{center}
\end{figure*}

\subsection{Multi-resolution Analysis}

As discussed earlier, 
one of the main advantages of kernel learning 
with relative comparisons is the feasibility of multi-resolution clustering using a single kernel matrix. 
To validate this claim, 
we repeat the {\em binary} and {\em multi-class} experiments described above. 
However, this time, we mix the binary and multi-class constraints and
use the same set of constraints in both experimental conditions.
We evaluate the results by performing binary and multi-class clustering, as before.

Figures~\ref{fig:binary_mixed} and \ref{fig:multiclass_mixed} 
illustrate the performance of different algorithms using the mixed set of constraints. 
Again, {\sf SKLR} produces more accurate clusterings, 
especially in the multi-class setting. 
In fact, two of the methods, {\sf SK$k$m} and {\sf ITML}, 
perform worse than the kernel $k$-means baseline in the multi-class setting.
On the other hand all methods outperform the baseline in the binary setting. 
The reason is that most of the constraints in the multi-class setting 
are also relevant to the binary setting, 
but the converse does not hold. Note that in the multi-class setting, {\sf sSKLR} fixes the early drop in the accuracy of {\sf SKLR} by handling the irrelevant constraints more efficiently.

Figure~\ref{fig:usps_vis} shows a visualization of the {\bf USPS Digits} dataset 
using the SNE method~\citep{sne} in the original space,
and the spaces induced by {\sf SK$k$m} and {\sf SKLR}.
We see that {\sf SKLR} provides an excellent separation of the clusters  
that correspond to even/odd digits as well as the sub-clusters that correspond to individual digits.

\subsection{Generalization Performance}

We now evaluate the generalization performance of the different methods 
to out-of-sample data on the {\bf MIT Scene} and {\bf USPS Digits} datasets
(recall that we do not subsample the {\bf Vehicles} dataset).
For the baseline kernel $k$-means algorithm, we run the algorithm on the whole datasets. 
For {\sf ITML} and {\sf LSML}, 
we apply the learned transformation matrix on the new out-of-sample data points. 
For {\sf SK$k$m}, {\sf SKLR}, and {\sf sSKLR},
we use Equation~(\ref{eq:out_of_sample}) 
to find the transformed kernel matrix of the whole datasets. 
The results of this experiment are shown in Figure~\ref{fig:generalization_results}.  
As can be seen from the figure, 
also in this case, 
when generalizing to out-of-sample data, 
{\sf SKLR} and {\sf sSKLR} produce significantly more accurate clusterings.

\begin{figure*}[t]
\setcounter{subfigure}{0}
	\begin{center}
\subfigure[]{\includegraphics[width=0.32\textwidth]{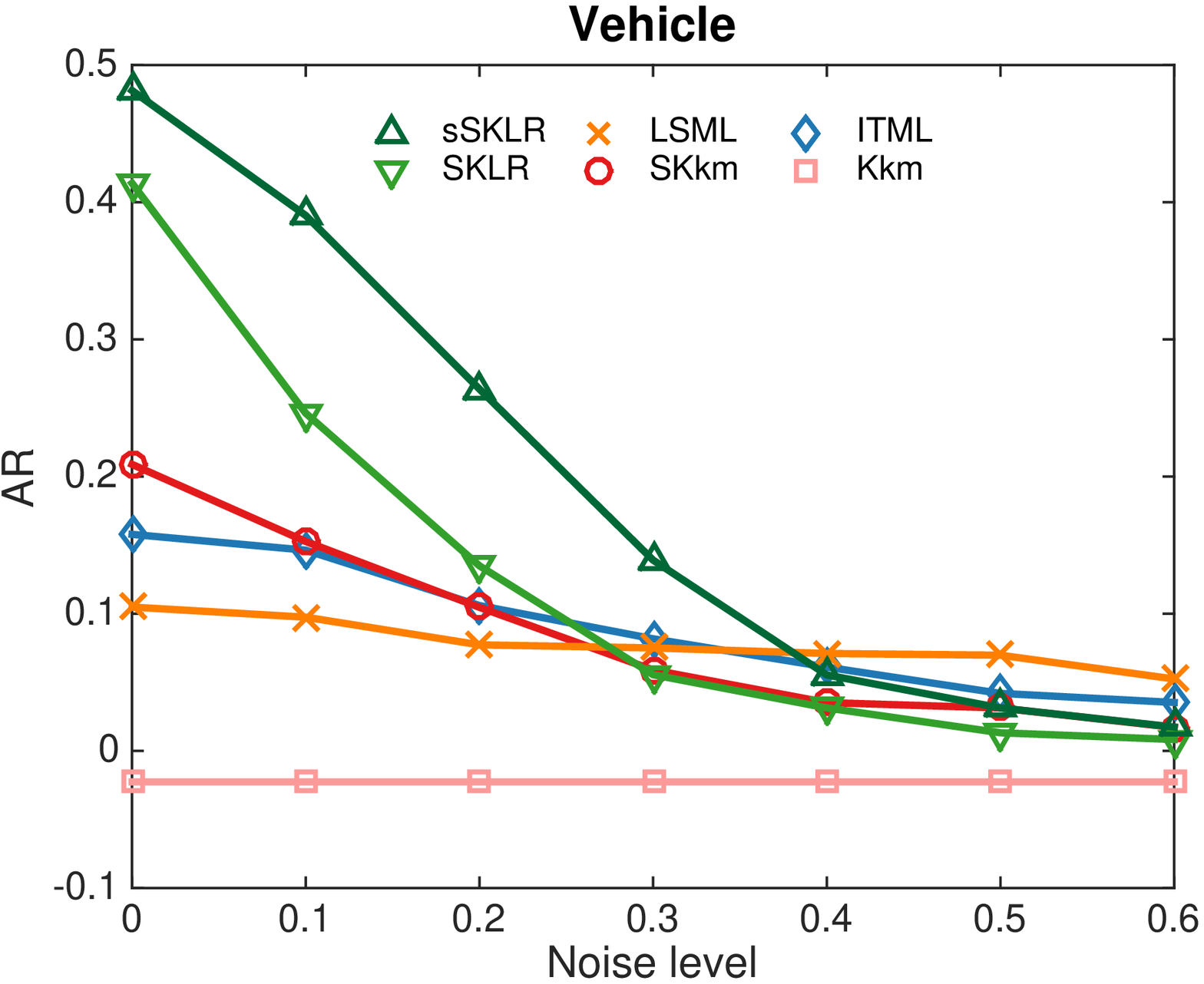}}
\subfigure[]{\includegraphics[width=0.32\textwidth]{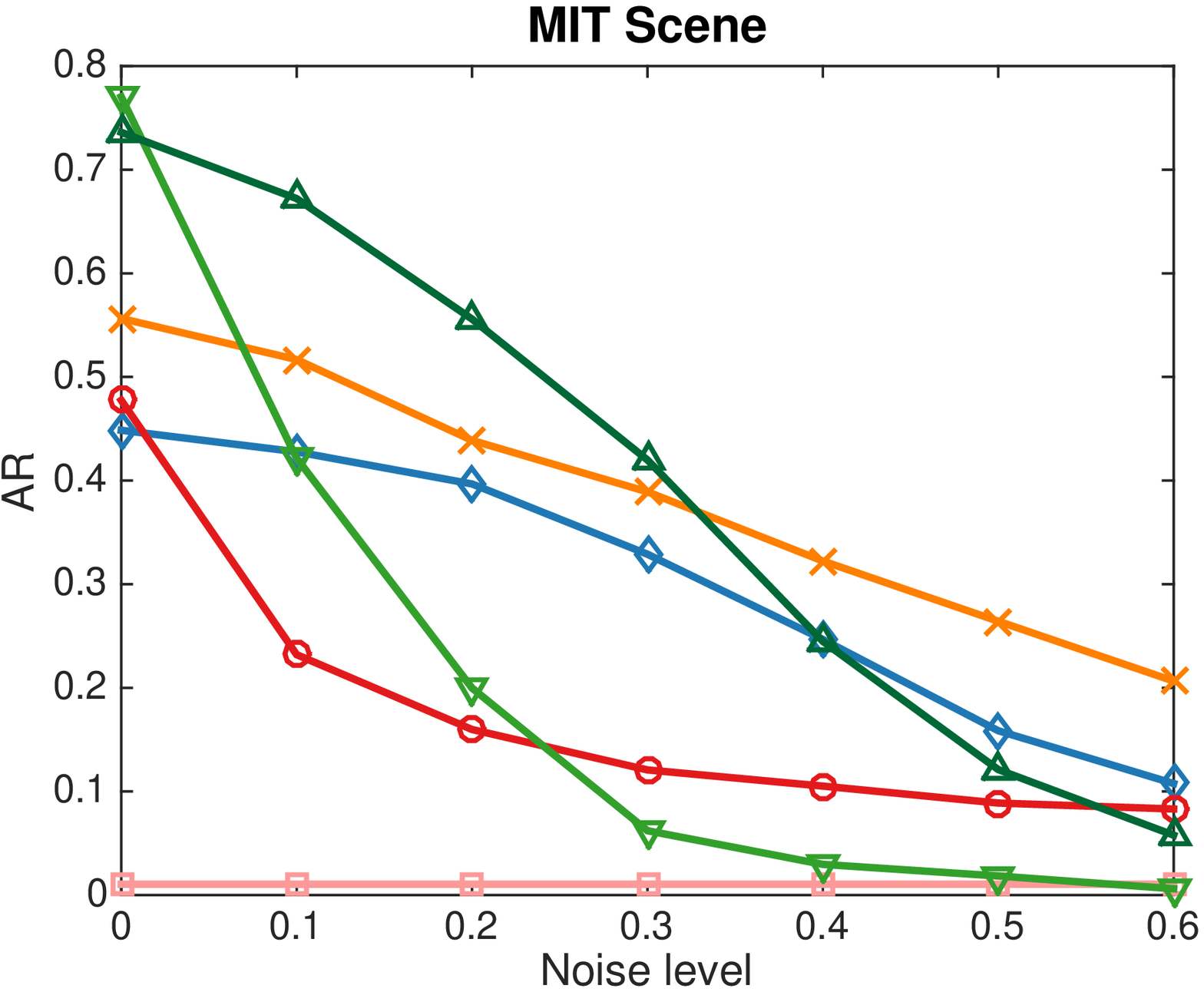}}
\subfigure[]{\includegraphics[width=0.32\textwidth]{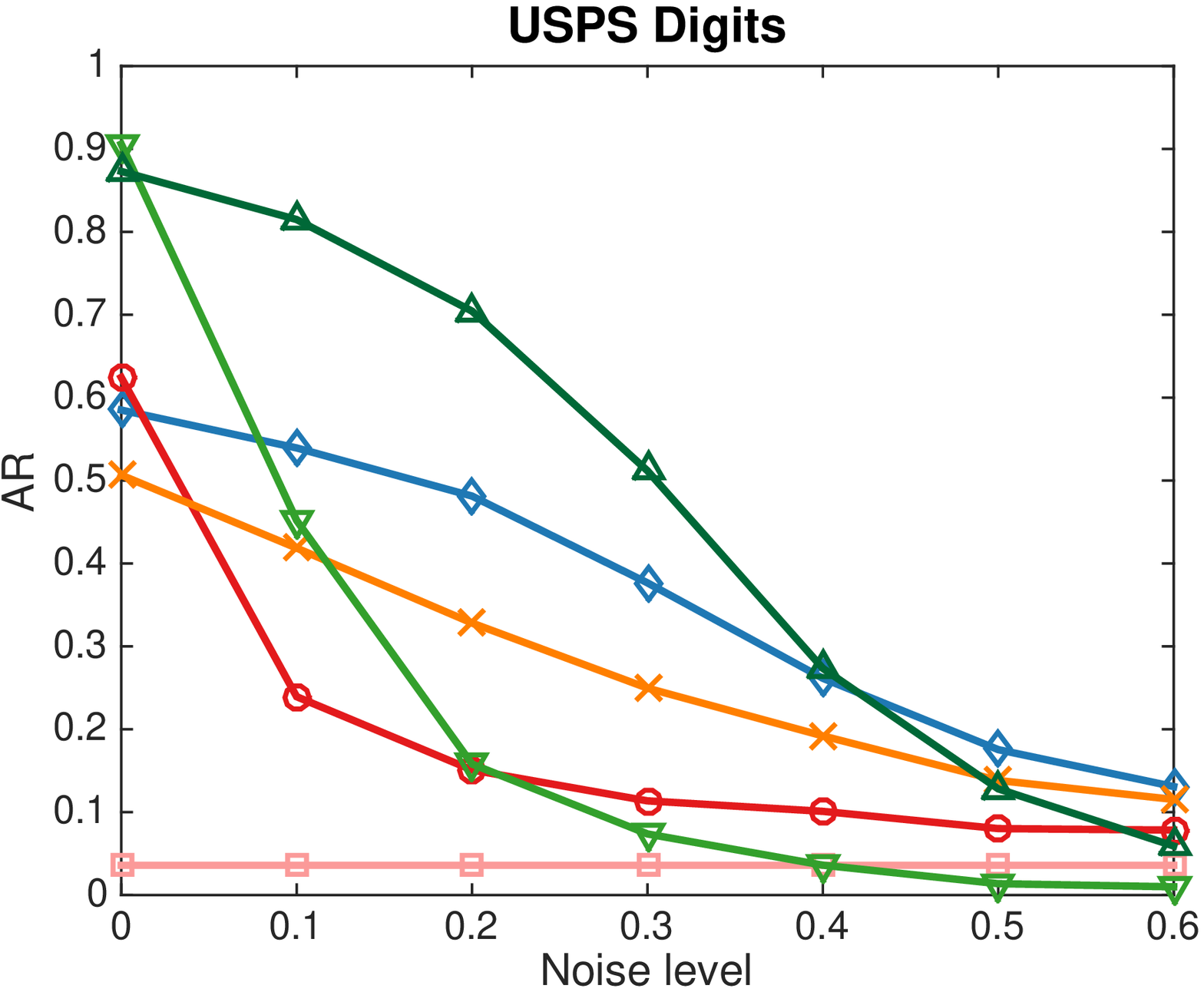}}
\caption{Effect of noise on the three datasets: (a)~{\bf Vehicle};
(b)~{\bf MIT Scene};
(c)~{\bf USPS Digits}.
}
\label{fig:noise_results}
\end{center}
\end{figure*} 

\subsection{Effect of Noise}
\label{sect:noise}

To evaluate the effect of noise, we first generate a fixed number of pairwise or relative constraints for each method ($360$, $720$, and $900$ constraints for {\bf Vehicle},  {\bf MIT Scene}, and {\bf USPS Digits}, respectively) and then corrupt up to $0.6$ of the constraints with step size of $0.1$. We corrupt the ML and CL constraints by simply flipping the type of the constraints. For the relative constraints, we randomly swap the outlier with one of inlier points. The results on the different datasets are shown in Figure~\ref{fig:noise_results}. As can be seen, the performance of the {\sf SLKR} drops immediately as the noise is introduced. This behavior is expected as the method is not tailored to handle noise.  The {\sf sSKLR} method, on the other hand, degrades more gradually as the level of noise increases. Clearly, {\sf sSKLR} outperforms all other methods until around $0.4$ noise level and  performs comparably good as the other methods for higher levels of noise. 

\subsection{Effect of Equality Constraints}

\begin{figure*}[t]
\setcounter{subfigure}{0}
	\begin{center}
\subfigure[]{\includegraphics[width=0.32\textwidth]{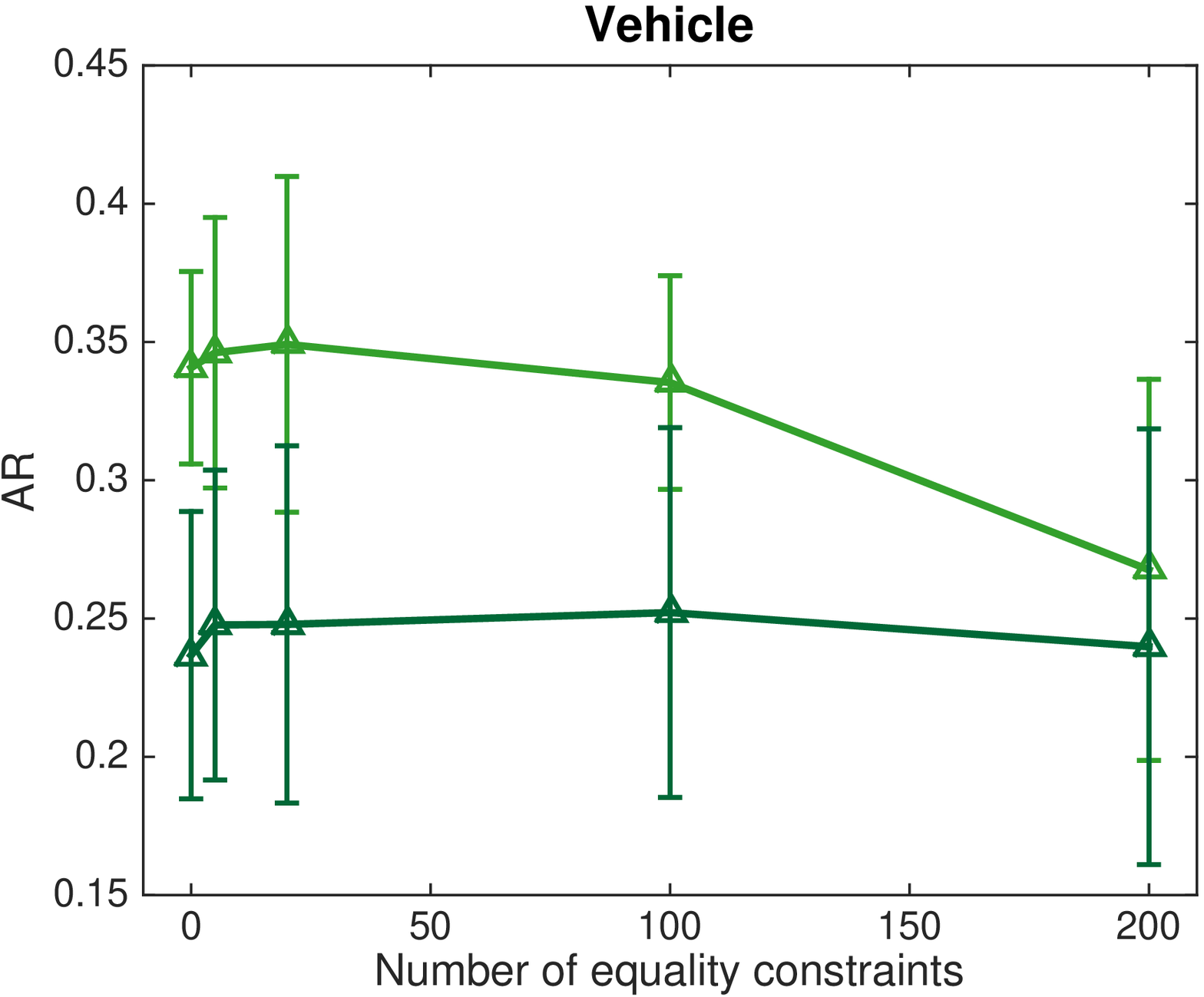}}
\subfigure[]{\includegraphics[width=0.32\textwidth]{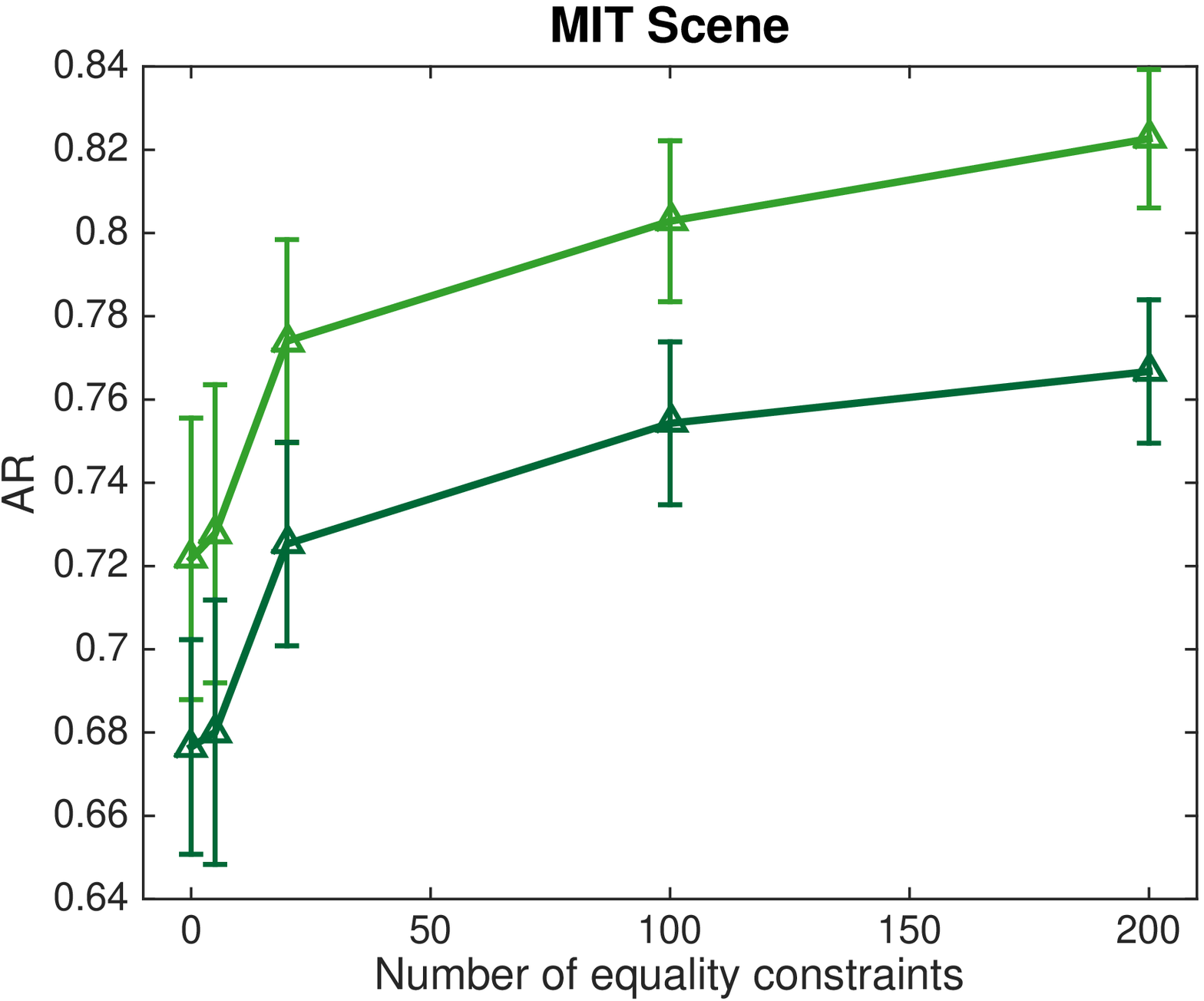}}
\subfigure[]{\includegraphics[width=0.32\textwidth]{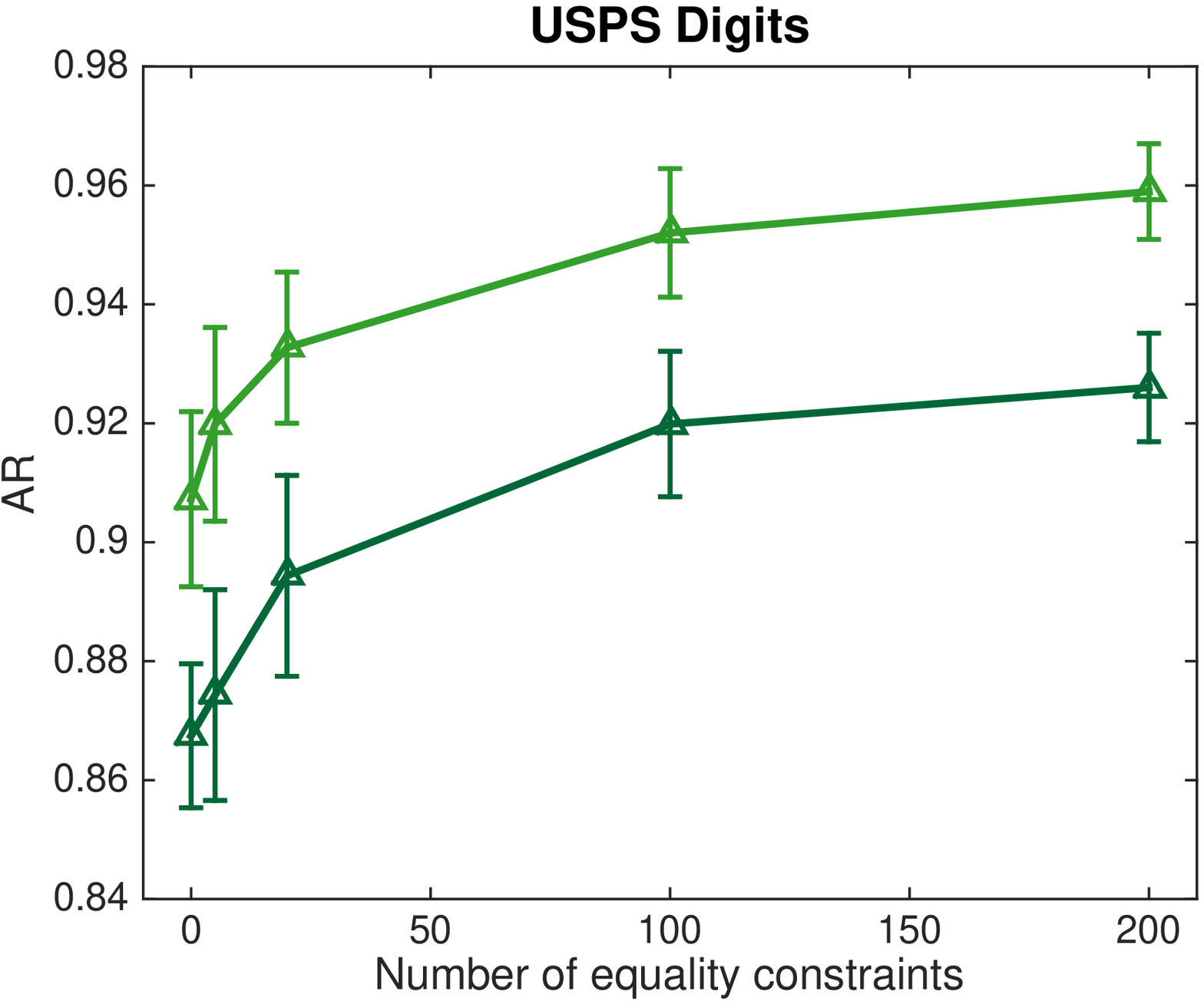}}
\caption{Effect of equality constraints on the three datasets: (a)~{\bf Vehicle};
(b)~{\bf MIT Scene};
(c)~{\bf USPS Digits}.
}
\label{fig:equality_results}
\end{center}
\end{figure*}

To evaluate the effect of equality constraints on the clustering, 
we consider a multi-class clustering scheme.
For all datasets, 
we first generate a fixed number of relative  comparisons ($360$, $720$, and $900$ relative comparisons for {\bf Vehicle},  {\bf MIT Scene}, and {\bf USPS Digits}, respectively)
and then we add some additional equality constraints (up to $200$). 
The equality constraints are generated by randomly selecting three data points, all from the same class, or each from a different class.
The results are shown in Figure~\ref{fig:equality_results}. 
As can be seen, considering the equality constraint also improves the performance, 
especially on the {\bf MIT Scene} and {\bf USPS Digits} datasets.
On the {\bf Vehicle} dataset and using the {\sf SKLR} method, the performance starts to drop after around $20$ constraints. The reason can be that after this point, many {\it unrelated} equality constraints, i.e., the ones with data points all coming from the same class, start to appear more and thus, reduce the performance. {\sf sSKLR} method, on the other hand, can handle the unrelated constraints.
Finally, note that none of the other methods that we consider can handle equality constraints.

\subsection{Running Time}

We note that in
in this paper, we do not 
report running-time results, since the implementations of the different methods are provided in different programming languages. 
However, all tested methods have comparable running times. 
In particular, the computational overhead of our method can be limited
by leveraging the fact that the algorithm has to work with low-rank $r\times r$ matrices. For all our datasets, we have $r \approx \frac{1}{2} n$. As an example, on a machine with a $2.6$ GHz processor and $16$ GB of RAM, performing $10$ passes over all constraints on the subset of $1000$ data points from the {\bf USPS Digits} dataset with number of constraints equal to $500$ takes around $670$ seconds. A higher speed-up can be obtained by further reducing the rank of the approximation (possibly sacrificing the performance).

\section{Conclusion}
\label{sec:conclusion}

We have devised a semi-supervised kernel-learning algorithm that
can incorporate relative distance constraints,
and used the resulting kernels for clustering.
In particular, our method incorporates relative distance constraints 
among triples of items, 
where labelers are asked to select one of the items as an outlier,
while they may provide a ``don't know'' answer.
The metric-learning problem is formulated as a kernel-learning problem, 
where the goal is to find the 
kernel matrix $\kernel$ that is the closest to an initial kernel $\kernel_0$
and satisfies the constraints induced by the relative distance constraints.
We have also introduced a soft formulation that can handle inconsistent constraints, 
and thus, reduce the robustness of the approach. 

Our experiments show that our method
outperforms by a large margin other competing methods, 
which either use ML/CL constraints
or use relative constraints but different metric-learning approaches.
Our method is compatible with
existing kernel-learning techniques~\citep{skms}
in the sense that if ML and CL constraints are available,
they can be used together with relative comparisons.

We have also proposed to
interpret an ``unsolved'' distance comparison
so that the interpoint distances are roughly equal.
Our experiments suggest that
incorporating such equality constraints to
the kernel learning task
can be advantageous,
especially in settings where
it is costly to collect constraints.


\appendix
\section*{Appendix A. Bregman Projection for the Rank-2 Inequality Constraint}
\label{sec:app-A}

We derive the dual of the optimization problem given in Equation~\eqref{eq:lagrange_form}

\begin{equation*}
 \kernel_{t+1} = 
 \underset{\kernel}{\text{arg\,min }} \logdet(\kernel,\kernel_t) + \alpha \tr(\kernel \Cmat).
 \end{equation*}
 
 Setting the derivative of above to zero yields the following update for $\kernel_{t+1}$
 \begin{equation*}
\kernel_{t+1}  = (\kernel_{t}^{-1} + \alpha \Cmat)^{-1}.
\end{equation*}
Now, substituting for $\kernel$ and using~\eqref{eq:logdet}, we have the dual problem
\begin{align*}
& \tr((\kernel_{t}^{-1} + \alpha \Cmat)^{-1}\, \kernel_t^{-1}) - \log \det((\kernel_{t}^{-1} + \alpha \Cmat)^{-1}\,\kernel_t^{-1} ) - n + \alpha \tr((\kernel_{t}^{-1} + \alpha \Cmat)^{-1}\, \Cmat)\\
= & \tr((\kernel_{t}^{-1} + \alpha \Cmat)^{-1}\, (\kernel_t^{-1} + \alpha\Cmat)) - \log \det((\kernel_{t}^{-1} + \alpha \Cmat)^{-1}\,\kernel_t^{-1} ) - n\\
= & n - \log \det((\kernel_{t}^{-1} + \alpha \Cmat)^{-1}\,\kernel_t^{-1} ) - n\\
= & \log \det(\kernel_t\, (\kernel_{t}^{-1} + \alpha \Cmat))\\
= & \log \det(\mathbf{I}_n + \alpha\kernel_{t} \Cmat))
\end{align*}
which is maximized w.r.t. $\alpha$.

\section*{Appendix B. Bregman Projection for the Soft Margin Rank-2 Inequality Constraint}
\label{sec:app-B}

Similarly, for the soft margin case, we have the Lagrangian form
\begin{equation*}
 \logdet(\kernel,\kernel_t) + \frac{1}{2}\,\lambda\, \xi^2 + \alpha\, \left(\tr(\kernel^\top \Cmat)-\xi\right)
 \end{equation*} 
We substitute for $\kernel$ as before and we set $\xi = \alpha/\lambda$ for the slack variable. This yields the following dual maximization problem w.r.t. $\alpha$
\begin{alignat*}{2}
& \tr((\kernel_{t}^{-1} + \alpha \Cmat)^{-1}\, \kernel_t^{-1}) - \log \det((\kernel_{t}^{-1} + \alpha \Cmat)^{-1}\,\kernel_t^{-1} ) - n && \\
 &\hspace{4.7cm}  + \alpha( \tr((\kernel_{t}^{-1} + \alpha \Cmat)^{-1}\, \Cmat) - \frac{\alpha}{\lambda}) && + \frac{1}{2} \lambda (\frac{\alpha}{\lambda})^2\\
 = & \tr((\kernel_{t}^{-1} + \alpha \Cmat)^{-1}\, (\kernel_t^{-1} + \alpha\Cmat)) - \log \det((\kernel_{t}^{-1} + \alpha \Cmat)^{-1}\,\kernel_t^{-1} ) - n && - \frac{1}{2} \frac{\alpha^2}{\lambda} \\
= &- \log \det((\kernel_{t}^{-1} + \alpha \Cmat)^{-1}\,\kernel_t^{-1} ) -  \frac{1}{2} \frac{\alpha^2}{\lambda} && \\
= & \log \det(\mathbf{I}_n + \alpha\kernel_{t} \Cmat)) -  \frac{1}{2} \frac{\alpha^2}{\lambda} && 
\end{alignat*}

\section*{Appendix C. Cholesky Decomposition of Identity Plus Rank-2 Matrix}
\label{sec:app-C}

We provide an algorithm to calculate the Cholesky decomposition of symmetric identity plus rank-2 matrix of the form $\mathbf{A} = \mathbf{I}_n + \lambda_1 \mathbf{u\,v}^\top + \lambda_2 \mathbf{w\,z}^\top$. Note that in order for $\mathbf{A}$ to be symmetric, $\mathbf{w}$ and $\mathbf{z}$ must be linearly dependent on $\mathbf{u}$ and $\mathbf{v}$. However, for simplicity, we omit this dependence and derive the update in terms of all vectors. In order to find the lower-triangular matrix $\mathbf{L}$ such that $\mathbf{L\, L}^\top = \mathbf{A}$, we perform the decomposition
\begin{equation*}
\mathbf{A} = \begin{bmatrix}
\sqrt{\tau} & 0\\
\lambda_1 v_1 \mathbf{u}_{2:n} + \lambda_2 z_1 \mathbf{w}_{2:n} & \mathbf{I}_{n-1}
\end{bmatrix}
\begin{bmatrix}
1 & 0\\
0 & \tilde{\mathbf{A}}_{2,2}
\end{bmatrix}
\begin{bmatrix}
\sqrt{\tau} & \lambda_1 u_1 \mathbf{v}_{2:n}^\top + \lambda_2 w_1 \mathbf{z}_{2:n}^\top\\
0 & \mathbf{I}_{n-1}
\end{bmatrix}
\end{equation*}
where $\tau = 1 + \lambda_1 u_1 v_1 + \lambda_2 w_1 z_1$. We can also write $\tilde{\mathbf{A}}_{2,2}$ in a diagonal plus rank-2 update form as follows.
\begin{equation*}
\tilde{\mathbf{A}}_{2,2} = \mathbf{I}_{n-1} + \lambda_1' \mathbf{u}_{2:n} (\mathbf{v}_{2:n} + \lambda_2 w_1 (z_1\mathbf{v}_{2:n} - v_1\mathbf{z}_{2:n}))^\top + \lambda_2' \mathbf{w}_{2:n} (\mathbf{z}_{2:n} - \lambda_1 u_1 (z_1\mathbf{v}_{2:n} - v_1\mathbf{z}_{2:n}))^\top
\end{equation*}
in which, $\lambda_1' = \lambda_1/\tau$ and $\lambda_2' = \lambda_2/\tau$. Thus, we can recursively apply the above procedure and solve for the Cholesky decomposition of matrix $\tilde{\mathbf{A}}_{2,2}$.

\vskip 0.2in
\bibliographystyle{plainnat}
\bibliography{refs-short}

\end{document}